\documentclass{article}

\PassOptionsToPackage{numbers, compress}{natbib}

\usepackage[preprint]{neurips_2025}

\makeatletter
\renewcommand{\@noticestring}{}
\makeatother

\usepackage[utf8]{inputenc} 
\usepackage[T1]{fontenc}    
\usepackage{url}            
\usepackage{booktabs}       
\usepackage{amsfonts}       
\usepackage{amsmath}        
\usepackage{nicefrac}       
\usepackage{microtype}      
\usepackage{xcolor}         
\usepackage{graphicx}       
\usepackage{datatool} 
\usepackage{array}
\usepackage{tabularx,multirow,booktabs}
\usepackage{hyperref}       
\usepackage{multirow}
\usepackage{makecell}
\usepackage{float}
\usepackage{graphicx}
\usepackage{subcaption}
\usepackage{nameref}
\usepackage{threeparttable}

\usepackage{listings}
\definecolor{ffgreen}{HTML}{3BA595}
\definecolor{ffgray}{HTML}{616C8C}
\definecolor{fforange}{HTML}{D95F18}

\lstdefinelanguage{CUDA}{
  language=C++,
  morekeywords=[1]{auto,const,typename,constexpr,noexcept,static_cast,dynamic_cast,template},
  keywordstyle=[1]\color{ffgray}\bfseries,
  morekeywords=[2]{__global__,__device__,__host__,__shared__,__constant__,__syncthreads,threadIdx,blockIdx,blockDim,gridDim},
  keywordstyle=[2]\color{ffgreen}\bfseries,
  morekeywords=[3]{torch,torch::Tensor,torch::IntArrayRef,torch::max_pool2d,.cpu(),.item},
  keywordstyle=[3]\color{fforange}\bfseries,
  identifierstyle=\color{black},
  commentstyle=\color{ffgray}\ttfamily,
  showstringspaces=false,
  breaklines=true,
  backgroundcolor=\color{gray!10},
  basicstyle=\ttfamily\footnotesize,
  frame=none,
  frameround=ffff,
  rulesepcolor=\color{gray},
  breakatwhitespace=false,
}

\lstdefinestyle{pythoncode}{
  language=Python,
  frame=lines,
  backgroundcolor=\color{gray!10},
  basicstyle=\ttfamily\scriptsize,  
  keywordstyle=\color{blue},
  commentstyle=\color{gray},
  stringstyle=\color{teal},
  showstringspaces=false,
  breaklines=true,
  numbers=none,                     
}
\usepackage{multicol}

\title{The FM Agent}

\author{
  Annan Li\thanks{core contributor},  
  Chufan Wu\footnotemark[1], 
  Zengle Ge\footnotemark[1],  
  Yee Hin Chong\footnotemark[1], 
  Zhinan Hou\footnotemark[1], 
  Lizhe Cao\footnotemark[1], \\
  \textbf{Cheng Ju}\footnotemark[1], 
  \textbf{Jianmin Wu}\footnotemark[1],
  \textbf{Huaiming Li, 
  Haobo Zhang, 
  Shenghao Feng, }\\
  \textbf{Mo Zhao, 
  Fengzhi Qiu, 
  Rui Yang,
  Mengmeng Zhang, 
  Wenyi Zhu, }\\
  \textbf{Yingying Sun, 
  Quan Sun, 
  Shunhao Yan,
  Danyu Liu,
  Dawei Yin\thanks{project sponsor}, 
  Dou Shen\footnotemark[2].}
  \And
  FM Agent Team, Baidu AI Cloud \\
}

\begin{document}

\maketitle

\begin{abstract}
Large language models (LLMs) are catalyzing the development of autonomous AI research agents for scientific and engineering discovery. We present FM Agent, a novel and general-purpose multi-agent framework that leverages a synergistic combination of LLM-based reasoning and large-scale evolutionary search to address complex real-world challenges. The core of FM Agent integrates several key innovations: 1) a cold-start initialization phase incorporating expert guidance, 2) a novel evolutionary sampling strategy for iterative optimization, 3) domain-specific evaluators that combine correctness, effectiveness, and LLM-supervised feedback, and 4) a distributed, asynchronous execution infrastructure built on Ray. Demonstrating broad applicability, our system has been evaluated across diverse domains, including operations research, machine learning, GPU kernel optimization, and classical mathematical problems. FM Agent reaches state-of-the-art results autonomously, without human interpretation or tuning — \textbf{1976.3} on ALE-Bench (+5.2\%), \textbf{43.56\%} on MLE-Bench (+4.0pp), up to \textbf{20×} speedups on KernelBench, and establishes new state-of-the-art(SOTA) results on several classical mathematical problems. Beyond academic benchmarks, FM Agent shows considerable promise for both large-scale enterprise R\&D workflows and fundamental scientific research, where it can accelerate innovation, automate complex discovery processes, and deliver substantial engineering and scientific advances with broader societal impact.

\begin{figure}[H]
    \centering
    \includegraphics[width=0.75\linewidth]{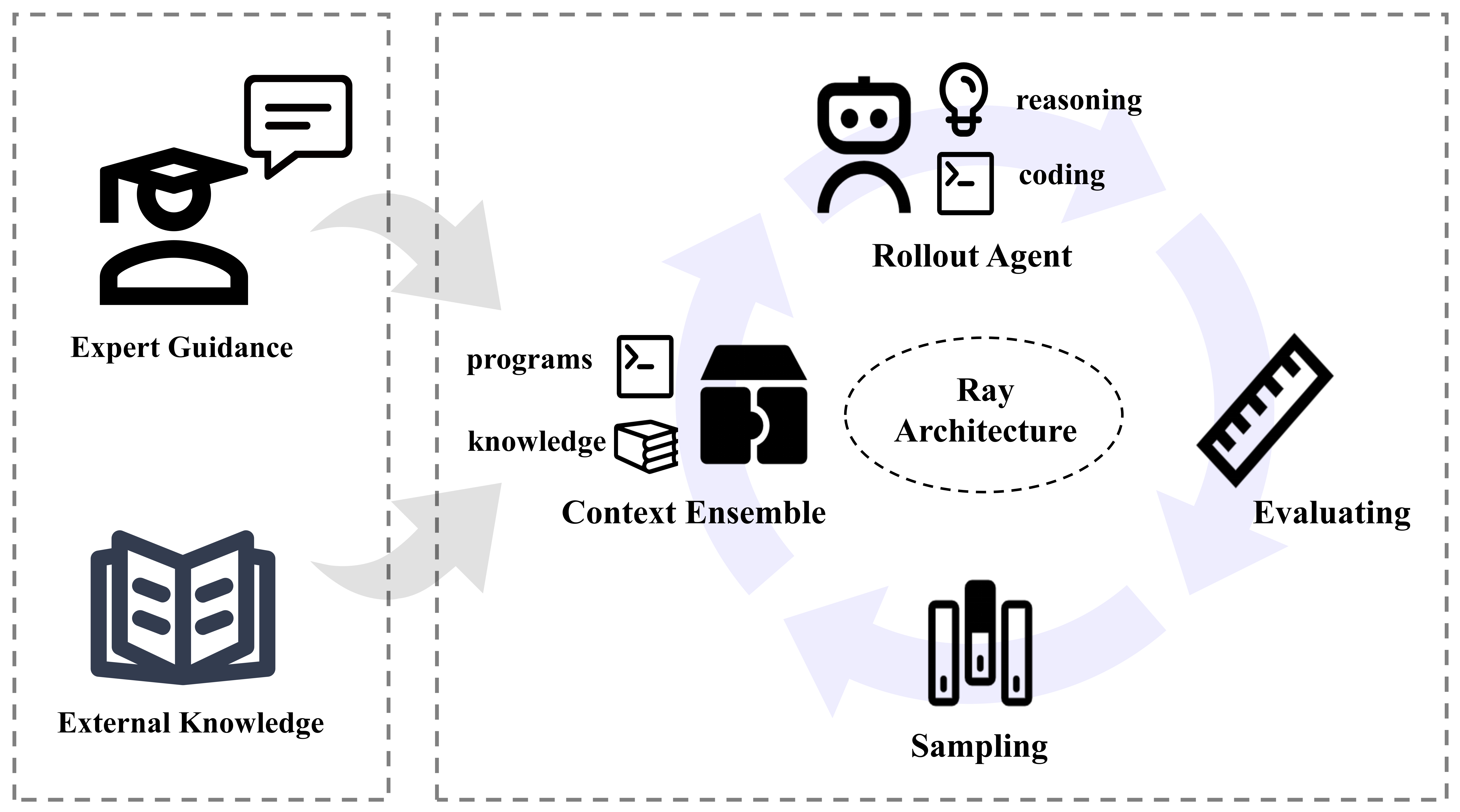}
    \caption{The workflow of FM Agent System to tackle a complex algorithm problem.}
    \label{fig:structure}
\end{figure}

\end{abstract}

\section{Introduction}
Recent advances in large language models (LLMs) have spurred the development of increasingly capable and autonomous AI research agents. A prominent line of work focuses on orchestrating multiple LLM-driven agents to tackle complex, open-ended discovery and optimization tasks. These systems often employ a search-driven paradigm, where agents explore solution spaces systematically through evolutionary or reinforcement-style loops. Pioneering systems demonstrate how LLMs can generate, mutate, and evaluate large populations of candidate solutions, enabling the discovery of novel and high-performing designs across various domains.

The applicability of such AI research agents extends far beyond academic benchmarks. In industrial settings, numerous high-impact challenges—from combinatorial optimization and time-series forecasting to high-performance kernel tuning—share a common structure: evaluating candidate solutions is relatively straightforward, whereas identifying truly effective ones is exceptionally difficult. Traditionally, progress in these areas has relied on specialized engineers who design specific algorithms and refine them through iterative, project-based optimization. This human-driven process intrinsically mirrors a research cycle: retrieving relevant knowledge, synthesizing ideas from diverse sources, and continuously testing and refining solutions. AI agents that embody large-scale search and evolutionary principles are particularly suited to automating this process, as they can maintain diverse candidate pools, apply intelligent variation operations, and leverage performance feedback to progressively evolve superior solutions.

To effectively harness these capabilities for real-world industrial problems, we propose FM Agent, a novel and general-purpose multi-agent framework. FM Agent is designed to be broadly applicable across domains such as operations research, machine learning, and system optimization. It integrates four key architectural innovations to achieve robust performance and scalability:

\begin{itemize}
\item \textbf{Cold-Start Initialization.} This phase integrates  diverse generation agents to produce a broad yet high-quality initial solution space. Moreover, with an optional expert-in-the-loop design, the framework ensures evolutionary search begins from a pragmatically grounded foundation, significantly accelerating convergence, especially in some real-world complex cases.

\item \textbf{Adaptive Diversity-Driven Sampling.} Our novel sampling strategy orchestrates multiple parallel evolutionary islands, adaptively balancing exploration and exploitation through dynamic resource allocation. This mechanism maintains productive diversity across algorithmic lineages while systematically steering the population toward global optima.

\item \textbf{Domain-Specific Evaluation.} Custom evaluators synthesize multiple critical criteria—including functional correctness, operational effectiveness, and LLM-supervised quality assessment—to generate nuanced, multi-faceted feedback. This comprehensive scoring mechanism provides rich, cumulative signals that precisely guide the iterative refinement process.

\item \textbf{Distributed Asynchronous Infrastructure.} Built on Ray, our scalable orchestration framework enables fine-grained, large-scale concurrent evaluation across distributed computing resources. This architecture ensures efficient resource utilization while facilitating rapid and systematic exploration of complex, high-dimensional solution spaces.
\end{itemize}

By unifying knowledge-augmented reasoning, autonomous evolution, and domain-aware evaluation within a scalable infrastructure, FM Agent constitutes a general self-improving system. It establishes new state-of-the-art results on authoritative benchmarks: achieving \textbf{1976.3} on ALE-Bench\cite{imajuku2025ale} (\textbf{+5.2\%}), \textbf{43.56\%} on MLE-Bench\cite{chan2024mlebench} (\textbf{+4.0pp}), and delivering \textbf{2.08$\times$ to 20.77$\times$} speedups over \texttt{torch.compile} on KernelBench\cite{kernelbench}. Furthermore, it demonstrates strong performance on classical mathematical problems and matches or surpasses real-world algorithmic practice in various industrial scenarios within and beyond Baidu. We believe FM Agent lays a foundation for a new generation of AI research agents capable of addressing tangible productivity challenges, thereby contributing to technological advancement and broader societal benefit.

\begin{figure*}[t]  
    \centering  
    \includegraphics[width=0.85\textwidth]{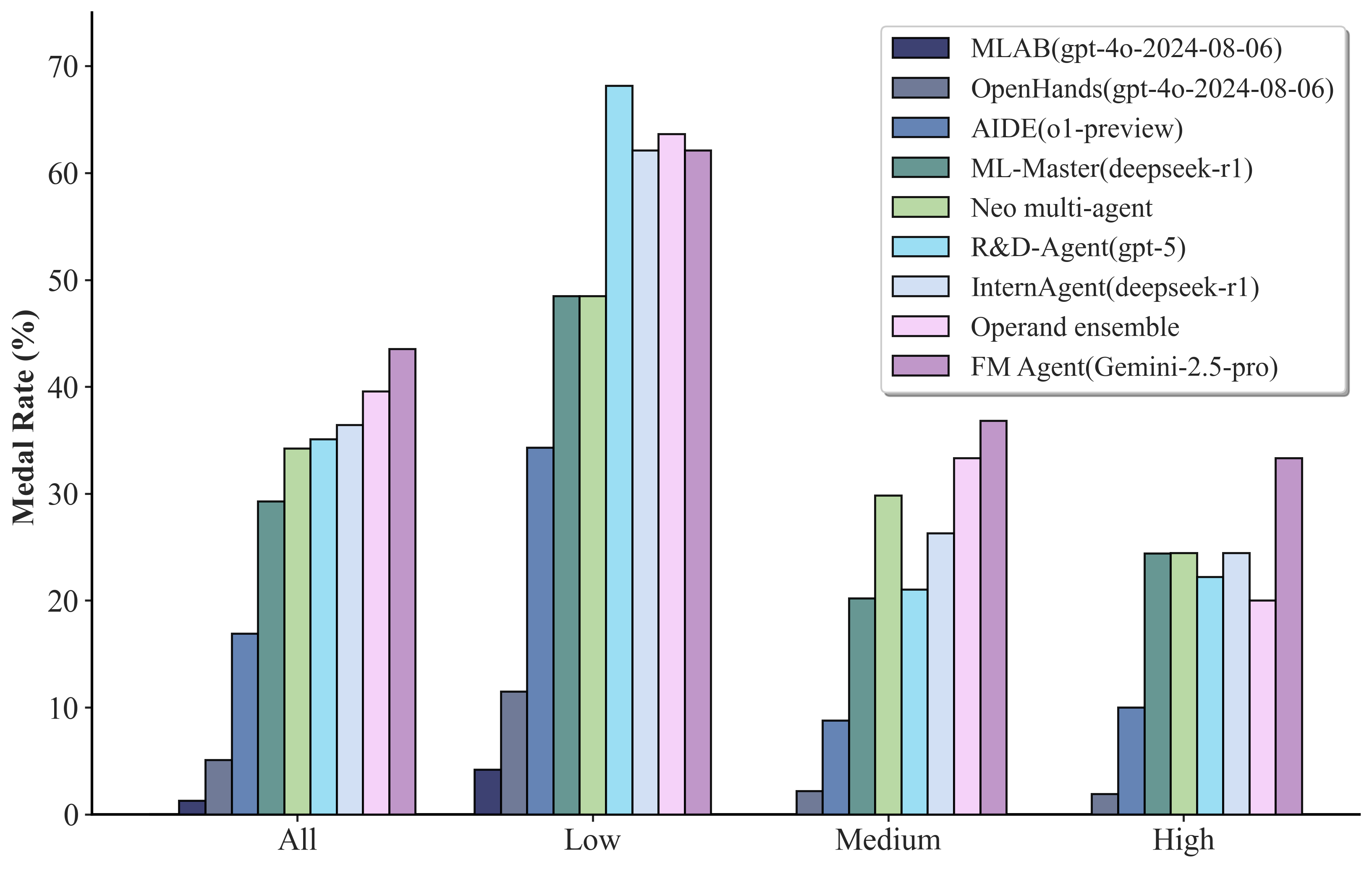} 
    \caption{Performance of Agents on MLE-Bench: Medal Rate (\%), evaluating FM Agent across real-world machine learning tasks sourced from Kaggle competitions.}
    \label{fig:mle} 
\end{figure*}


\begin{figure*}[t]
  \centering
  \includegraphics[width=0.85\textwidth]{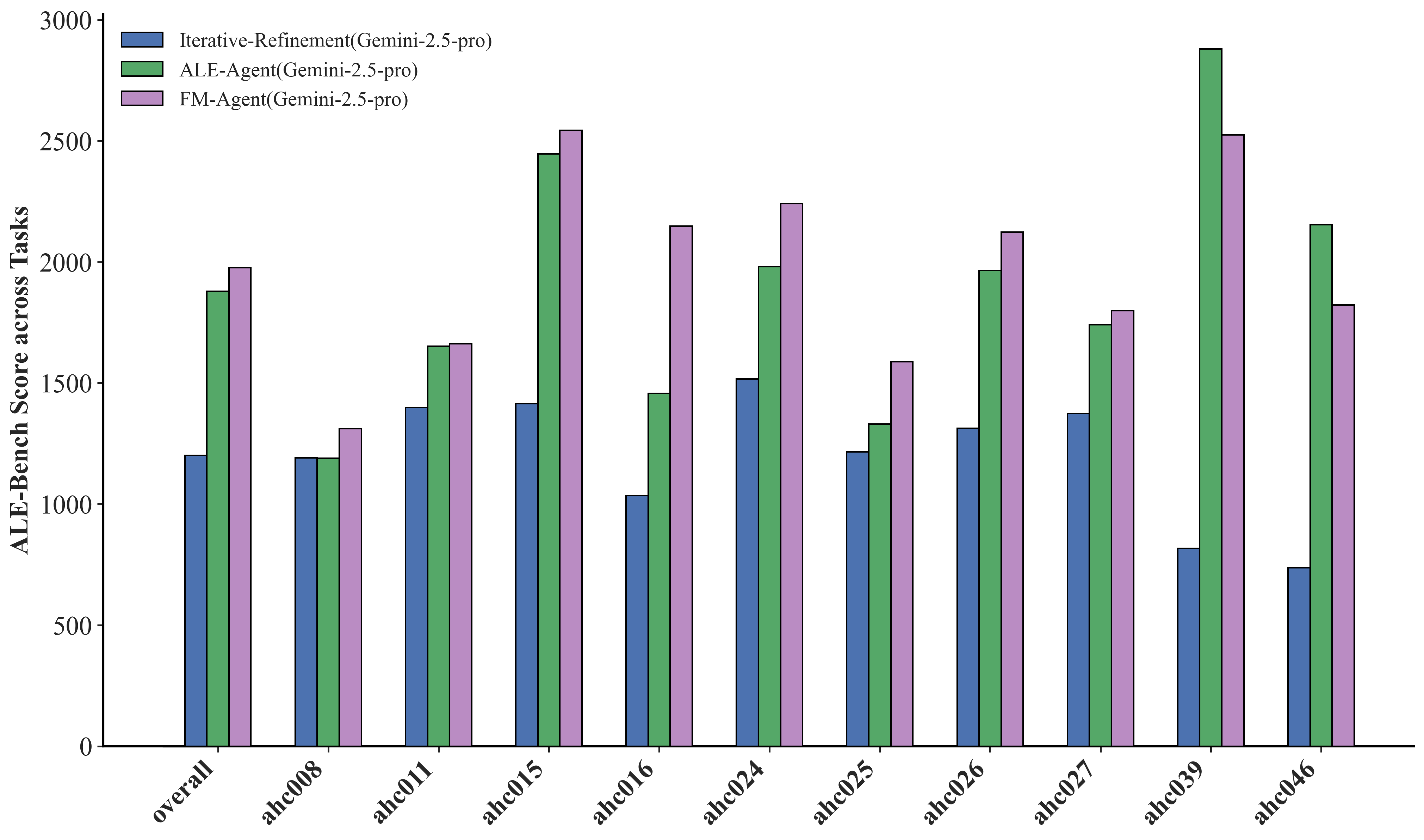}
  \caption{Performance of Agents on the ALE-Bench Lite, denoting the SOTA capability of FM Agent in tackling challenging heuristic-driven tasks from AtCoder Completion.}
  \label{fig:ale}
\end{figure*}

\section{Scenario}

\subsection{Machine Learning}
Machine Learning (ML), which enables computational models to learn patterns from data autonomously, has become fundamental to building intelligent systems, with critical applications including financial risk control \cite{li2025financial}, time series forecasting \cite{unknown2025hybrid}, and equipment fault diagnosis \cite{cho2025structured}.

However, developing high-performance ML models remains challenging due to the inherent complexity of data and optimization. Engineers often rely on iterative experimentation and object-specific customization in feature and model design, requiring substantial expertise and remains difficult to fully automate.

In this context, FM Agent emerges as a promising solution by integrating large language models with evolutionary computing. Moving beyond conventional automation boundaries that focus primarily on model selection and hyperparameter tuning, FM Agent facilitates autonomous orchestration of the complete ML workflow—from intelligent feature engineering to adaptive model construction. This approach demonstrates the potential to significantly reduce manual intervention while maintaining competitive performance in diverse tasks.

In our practice, FM Agent revolutionizes machine learning workflows through four principal approaches:
\begin{itemize}
\item \texttt{Autonomous Feature Mining:} Feature engineering is a crucial yet highly expertise-dependent and time-consuming aspect of machine learning. Guided by an evolutionary framework, FM Agent can autonomously explore raw data, surpassing the limitations of traditional statistical methods. Through iterative generation, evaluation, and filtering, it constructs informative new features from raw data, even discovering feature representations that are non-intuitive to human analysts yet possess high predictive power, thereby providing a superior data foundation for subsequent model learning.

\item \texttt{Intelligent Feature Combination:} The expressive capacity of individual features is limited, while manually creating effective feature interactions becomes exponentially complex with increasing dimensionality. FM Agent serves as an efficient "explorer" that systematically searches the high-dimensional feature combination space. It experiments with various mathematical operations and logical combinations to identify synergistic effects between features that maximize model performance. This process automates the critical step of discovering high-order nonlinear relationships, effectively enhancing the model's representational capability.

\item \texttt{Adaptive Model Fusion:} Model fusion (ensemble learning), a commonly used and effective technique in modern machine learning, combines multiple base learners to enhance predictive performance and robustness. FM Agent can train new  models for different purpose and develop more sophisticated fusion mechanisms beyond simple voting  strategies. For example, it can autonomously design weighted ensemble schemes, allocating appropriate weights to different sub-models based on their performance across various data subsets, or construct stacked ensemble models while optimizing the structure and parameters of the meta-learner, ultimately achieving predictive accuracy superior to any single component model.

\item \texttt{End-to-End Machine Learning Task Solving:} The most forward-looking application involves deploying FM Agent as an end-to-end machine learning system. In this paradigm, the agent receives a dataset and task objectives, then autonomously decides and executes the complete machine learning pipeline, including feature preprocessing, algorithm selection, model structure design, training, and validation. Through sequential decision-making guided by feedback from evolutionary cycles, it progressively builds an optimized machine learning pipeline tailored to specific tasks, significantly advancing the realization of fully automated machine learning.
\end{itemize}

In summary, these four directions clearly demonstrate FM Agent's evolution from a tool assisting specific tasks to a collaborative partner driving full-process automation. By leveraging its complex reasoning capabilities and integrating them with the systematic exploration of evolutionary frameworks, FM Agent has the potential to redefine the development efficiency and performance limits of machine learning solutions.

\subsection{Combinatorial Optimization}
Combinatorial optimization (CO), defined as the selection of optimal objects from finite solution sets, represents a foundational paradigm in operations research. This methodology formulates critical real-world applications including production scheduling \cite{floudas2005mixed}, logistics transportation \cite{demirel2016mixed}, and resource management \cite{cheng2003integrated}, where solving efficiency and solution quality directly translate to substantial economic value.

However, the NP-hard nature of CO problems renders exact solutions computationally prohibitive or intractable in practice. Consequently, significant research efforts focus on accelerating CO solving while preserving solution quality. Yet human-designed strategies often demand extensive domain expertise and costly trial-and-error processes \cite{achterberg2009scip}. FM Agent opens up a more transformative research frontier where agents actively participate in the discovery and refinement of new optimization strategies. Specifically, FM Agent is adopt into three principal avenues through which combinatorial optimization could be revolutionized: 
\begin{itemize}
    \item \texttt{Autonomous Design of Novel End-to-End Heuristics:} Heuristics are essential tools for tackling NP-hard CO problems, providing efficient, near-optimal solutions where exact methods fail. Guided by an evolutionary framework, FM Agent can explore the vast design space of heuristic algorithms. By iteratively generating, testing, and refining algorithmic components based on performance feedback, FM Agent can autonomously discover novel and powerful heuristics tailored to specific problem structures. This approach emulates the human process of innovation but at a scale and speed previously unattainable, potentially yielding strategies that diverge significantly from conventional designs.
    \item \texttt{Intelligent Augmentation of Optimization Solvers:}  Rather than replacing traditional solvers, FM Agent can serve as specialized collaborators to enhance their performance. Modern solvers for problems like mixed integer programming or constraint programming are modular systems that can be improved by integrating high-quality, problem-specific components. FM agent could be set to design these critical modules. For instance, it could help generate cutting planes to tighten linear relaxations, design presolve policy to accelerate solution process.
    \item \texttt{Direct and Iterative Solution Generation:}  Another significant application involves employing the FM Agent as a direct, end-to-end solver that iteratively constructs a high-quality solution. In this paradigm, FM Agent engages in a sequential decision-making process, where each step involves selecting a component of the solution. This method bypasses the need for an explicit intermediate algorithmic representation and instead enables the agent to develop a deep, implicit understanding of the problem's structure, reasoning its way directly to an optimal or near-optimal solution.
\end{itemize}

\subsection{Kernel Generation}

The explosive growth of deep neural networks (DNNs), particularly large-scale models such as large language models (LLMs), has established GPUs as the dominant platform for AI workloads. At the foundation of this stack lie CUDA kernels for DNN operators, which directly determine how effectively parallelism and memory hierarchies are utilized. Their efficiency is critical for overall system performance \cite{auto}, while inefficient kernels can significantly degrade model throughput.

Designing high-performance CUDA kernels is notoriously difficult due to the \textit{complex interactions} among memory access patterns, thread block configurations, and instruction scheduling \cite{gpuwc}. Manual tuning is often a labor-intensive trial-and-error process. Although AI compilers \cite{triton,tvm,torchdynamo} and domain-specific languages (DSLs) \cite{mirage,tilelang,mlir} provide automation for common operators, their reliance on predefined schedules and rigid transformation rules limits generalization to new workloads and prevents full exploitation of hardware-specific opportunities. This leaves a persistent performance gap.

Instead of treating kernel optimization as a human-driven, knowledge-intensive task, FM Agent reformulate it as an autonomous, data-driven process. Advances in LLMs have made scalable code generation possible \cite{vega,llm-vectorizer,cuasmrl}, but training them for high-quality performance is prohibitively expensive. FM Agent circumvents this by generating diverse candidates, evaluating their runtime performance, and using feedback to guide further exploration. This iterative loop transforms LLMs’ generative capacity into continuous, adaptive optimization. Crucially, scaling test-time computation yields increasingly specialized CUDA kernels.

\subsection{Math}

Beyond classical optimization, many mathematical tasks—such as theorem proving, inequality tightening, bound estimation, and geometric construction—can be reframed as search problems. Instead of seeking explicit optimal solutions, these tasks aim to uncover tighter analytical bounds or constructive proofs that approximate theoretical optima \cite{gonccalves2017hermite,novikov2025alphaevolve}. Recent theorem-proving systems, though highly advanced, typically depend on human-in-the-loop reasoning and rigid symbolic strategies, which may constrain scalability and exploration depth \cite{chen2025seed,huang2025gemini}.

FM Agent offers a approach by combining symbolic reasoning, numerical experimentation, and evolutionary exploration within a unified framework. It can iteratively generate alternative search strategies, construct target objects, and select relevant prior knowledge, guided by feedback on correctness, efficiency, and a deeper understanding of the current solution. Through such adaptive refinement, FM Agent can autonomously discover near-optimal solutions—achieving better performance criteria and occasionally revealing unexpected theoretical insights.

This perspective treats mathematics as an open-ended search landscape, where reasoning and evolution jointly drive discovery, enabling agents to assist in deriving stronger results and accelerating theoretical innovation.

\section{Framework}
The framework of FM Agent is designed as a two-stage process to autonomously discover and refine solutions for complex problems. In the Cold Start Stage, several generative agents are applied to solve the problem, aiming to rapidly generate a diverse pool of high-quality algorithms by learning from feedback and acting on instructions. In the subsequent phase, the generated algorithms are partitioned according to the maximum similarity between islands. During this process, we define a number of clusters equal to the number of islands and assign a set of clusters to each island, thereby initiating the subsequent Evolve Stage. In Evolve Stage, an evolutionary search is applied to innovate and improve upon these initial solutions through mutation and crossover mechanisms of island-based population. To achieve high evolutionary efficiency at scale, the framework is deployed on a high-performance distributed cluster that supports large-scale parallel execution. This design significantly enhances throughput, scalability, and overall convergence speed during the evolutionary process.

\begin{figure}[H]
    \centering
    \includegraphics[width=\linewidth]{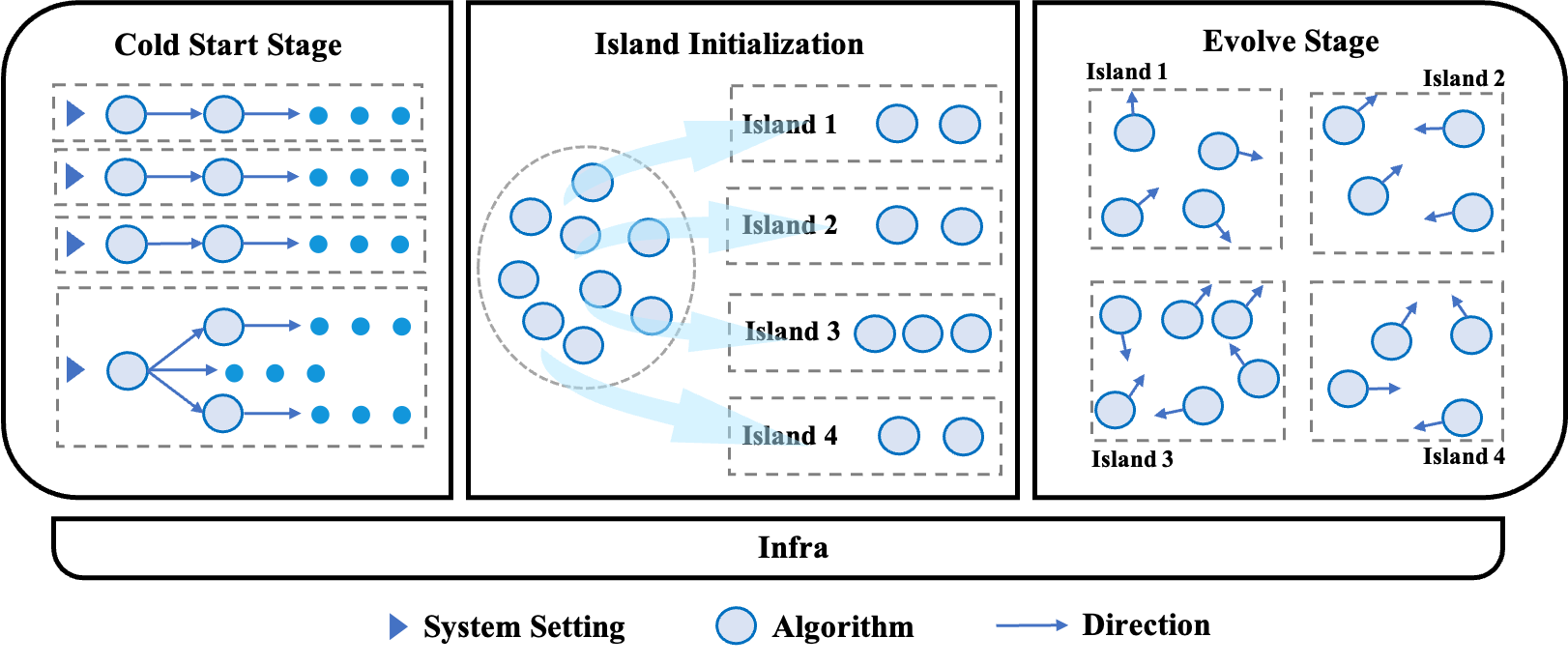}
    \caption{Framework of FM Agent with Cold Start Stage and Evolve Stage, both account for the final performance.}
    \label{fig:process}
\end{figure}

\subsection{Cold Start Stage}
The Cold Start Stage is dedicated to constructing an initial population of solutions with high diversity, thereby expanding the global solution space and laying a robust foundation for subsequent evolution, effectively mitigating the risk of premature convergence.

\textbf{Multi-Agent Parallel Expansion}. The system integrates diverse types of agents, supporting synchronous exploration of different generation strategies and optimization directions through differentiated configuration and parallel execution. This design significantly enhances the diversity of the initial population and provides the system with excellent scalability.

\textbf{Proactive Solution Space Expansion}. By guiding agents to explore divergent regions of the objective space, the system consciously broadens the coverage of the potential solution space during initialization. This approach reduces the risk of the evolutionary process becoming trapped in local optima and creates favorable conditions for subsequent in-depth optimization.

\subsection{Evolve Stage}
The Evolve module implements the core logic of FM Agent, orchestrating a large-scale, population-based search to innovate and improve upon the initial solutions. Its design is centered on principles of diversity preservation, adaptive evolution, and multi-population dynamics, which are encapsulated in an Efficiency Evolution Strategy.

\textbf{Multi-Population Island Model}. FM Agent utilizes a multi-population approach, where solutions are segregated into parallel "islands". On the one hand, each island evolves its population independently in the most time, allowing FM Agent to explore distinct regions of the solution space simultaneously and maintain diverse algorithmic families. On the other hand, the framework also facilitates periodic interaction between these islands, promoting cross-pollination of ideas and preventing the overall search from stagnating in local optima.

\textbf{Adaptive Diversity-Driven Sampling}. The framework employs an adaptive control system to steer evolution within each island, emphasizing diversity preservation through semantic and structural metrics to avoid premature convergence. A novel cluster-based sampling strategy is adopted, which adaptively maintains a balance between exploration and exploitation by dynamically adjusting selective pressure according to real-time population diversity. Moreover, a curated elite pool retains top-performing solutions to guide future generations. The whole adaptive mechanism ensures both sustained innovation and efficient convergence across evolutionary islands.



\textbf{Domain-Specific Evaluator}.
To address diverse evaluation demands, the framework employs a flexible, multi-faceted evaluation module providing both general and specialized feedback. General methods include a traditional single fitness score for quantitative ranking and LLM judge feedback for nuanced, qualitative assessment. For complex scenarios, such as machine learning or kernel generation, advanced domain-specific strategies assess multi-dimensional performance. These utilize special fitness metrics that are both numerical (e.g., balancing accuracy and latency) and contextual (e.g., resource utilization), ensuring the evolutionary process is guided by comprehensive domain requirements.


\begin{figure}[H]
    \centering
    \includegraphics[width=0.6\linewidth]{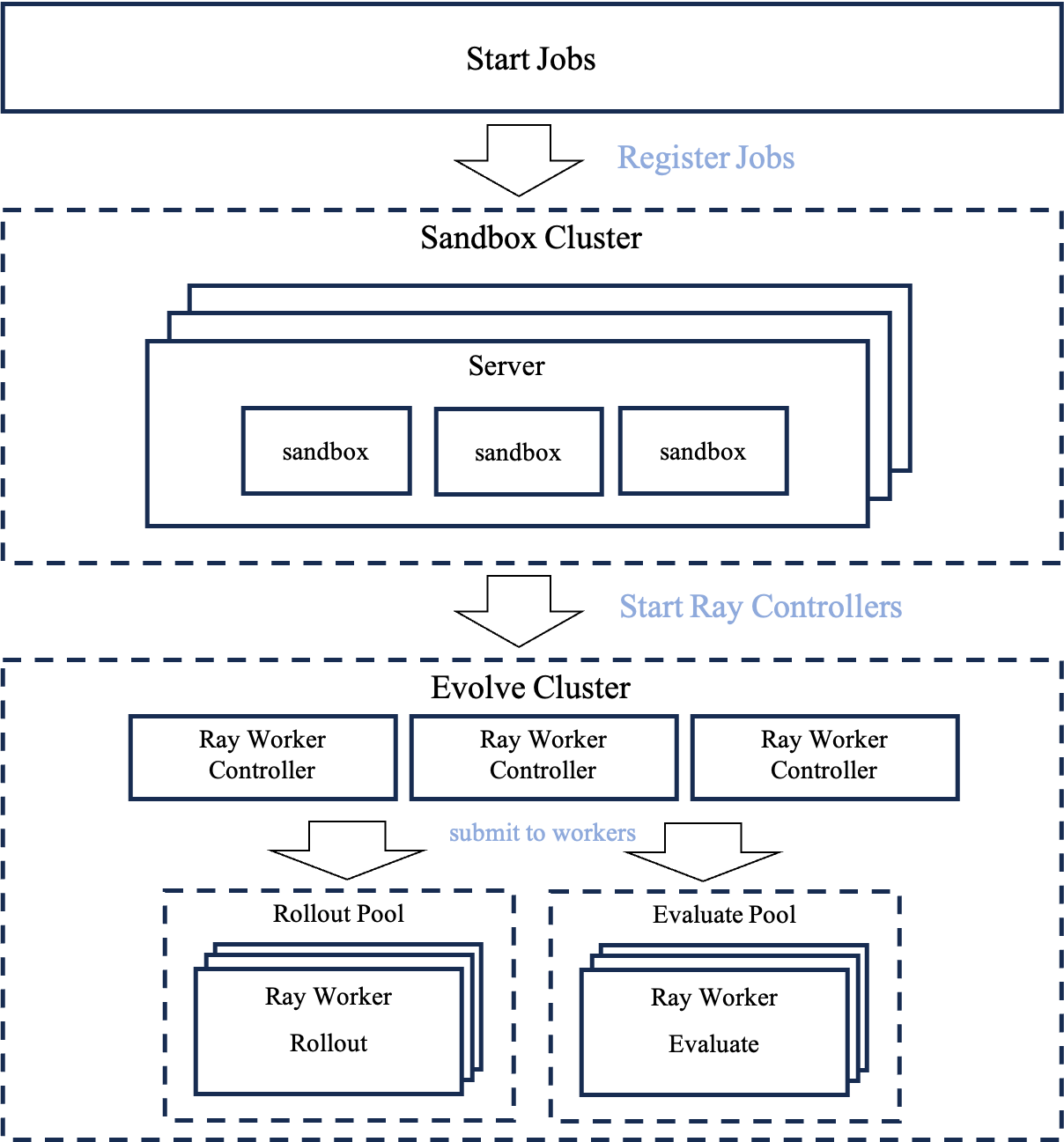}
    \caption{Architecture of the Large-Scale Distributed Evolutionary Cluster.}
    \label{fig:cluster_structure}
\end{figure}


\subsection{Infra}
FM Agent operates on a scalable distributed infrastructure purpose-built for high-throughput evolutionary computation. Its design emphasizes two key principles:

\textbf{Distributed Architecture}.
The system is orchestrated by the Ray framework, enabling seamless scaling from single-node to large multi-node clusters. Tasks are distributed across independent worker nodes, supporting concurrent execution of thousands of evolutionary processes with efficient resource utilization and fault tolerance.

\textbf{Asynchronous Generation–Evaluation Pipeline}.
The two primary workloads, including program generation and program evaluation, are executed in separate, parallelized worker pools. This separation allows asynchronous scheduling and dynamic load balancing, ensuring that compute-intensive synthesis and evaluation phases do not block each other. The result is significantly improved throughput, stability, and overall evolutionary efficiency.

\subsection{Human-Interactive Feedback Module}
An optional but recommended Human-Interactive Feedback Module is designed to flexibly incorporate domain expertise into the autonomous evolution process.
It centers on two key capabilities — real-time interaction and knowledge enhancement — enabling experts to guide and enrich the system without breaking its autonomy.

\textbf{Real-Time Monitoring and Interactive Intervention}. The system provides a panoramic visual interface for monitoring the evolution process, allowing experts to track key metrics such as fitness changes and population diversity in real time. Through natural language instructions (e.g., "prioritize model inference efficiency") or code-level interventions, experts can directly steer the evolutionary direction, ensuring alignment between optimization goals and business objectives.

\textbf{Knowledge Base Integration and Retrieval Augmentation}. The system supports the construction and maintenance of an expert knowledge base. By leveraging RAG technology, it enables efficient retrieval of structured knowledge, such as domain literature and best practices. When specific optimization challenges are encountered, the system automatically retrieves relevant knowledge fragments to inform mutation and crossover operations, thereby enhancing the rationality and interpretability of the search process.

\section{Experiment}
The efficacy and generalization capabilities of FM Agent were empirically evaluated through a battery of experiments conducted across three distinct yet computationally demanding domains. This evaluation was performed using a suite of established benchmarks, each selected to probe a core competency of the agent's autonomous, self-evolutionary framework. The chosen benchmarks were ALE-Bench, to assess ability for solving combinatorial optimization problems; MLE-Bench, to evaluate competency in automating complex, real-world machine learning engineering workflows ; and KernelBench, to quantify proficiency in the specialized task of generating high-performance GPU kernels. The results indicate that FM Agent establishes new state-of-the-art performance on these benchmarks, thereby demonstrating the robustness and generalizability of its problem-solving architecture. It is important to note that for all three benchmarks, optimization was performed exclusively by the LLM without human intervention.

\subsection{MLE-Bench}
MLE-Bench \cite{chan2024mlebench}, introduced by OpenAI, is an extensive benchmark created to assess systems on complex real-world machine learning tasks based on Kaggle competitions. MLE-Bench comprises 75 machine learning engineering–related competitions from Kaggle, forming a diverse collection of challenging tasks that evaluate practical ML engineering abilities, including model training, dataset processing, and experiment execution. 

We evaluate FM Agent on complete MLE-Bench among 75 machine learning tasks. We use the same evaluation metric as MLE-Bench. The results are shown in Table 1 and demonstrate that:

\begin{itemize}
\item \textbf{Submission Reliability and Broad Competence}: FM Agent demonstrates exceptional robustness by achieving valid submissions in \textbf{96.89\%} of tasks, a rate that matches or exceeds all other benchmarked agents. This near-perfect submission success rate underscores its remarkable capability to handle a wide array of machine learning challenges effectively and reliably, establishing a consistently high baseline performance.
\item \textbf{Superior Performance Against Human Benchmark}: The agent's results are particularly notable when compared to human performance. It surpasses more than half of all human submissions (Above Median) in \textbf{51.56\%} of the tasks, significantly outperforming other advanced agents such as InternAgent (48.44\%) and ML-Master (44.9\%). This indicates a strong and generalizable problem-solving ability that is highly competitive within the community.
\item \textbf{High Performance Ceiling and Medal Attainment}: FM Agent achieves the highest rate of medal acquisition (Any Medal) among all evaluated systems at \textbf{43.56\%}, with a standout performance in securing Gold medals in \textbf{22.67\%} of tasks. This exceptional medal distribution, especially the leading gold medal rate, highlights a superior performance ceiling and suggests that the agent can exceed the capabilities of the majority of human machine learning researchers in specific, complex task scenarios.
\end{itemize}

\begin{table}[h]
    \caption{FM Agent surpasses all baseline models across every evaluation dimension defined in MLE-Bench. All values are in percentage (\%). The results for MLAB(gpt-4o-2024-08-06), OpenHands(gpt-4o-2024-08-06), AIDE(o1-preview), R\&D-Agent(gpt-5), ML-Master(deepseek-r1), Neo multi-agent, InterAgent(deepseek-r1) and Operand ensemble(gpt-5, low verbosity/effort) are taken from the official MLE-Bench report. Results for FM Agent are averaged over three independent runs with different random seeds and are presented as the mean ± one standard error of the mean (SEM). The best-performing model in each category is highlighted in bold.}
    \label{tab:mle-leaderboard}
    \centering 
    \setlength{\heavyrulewidth}{1.5pt}  
    \setlength{\lightrulewidth}{0.5pt}   
    \resizebox{\linewidth}{!}{
        \begin{tabular}{@{}lcccccc@{}}  
            \toprule  
            \hspace{0.5em}\textbf{Agent} &  
            \textbf{\makecell{Valid \\ Submission}} &  
            \textbf{\makecell{Above \\ Median}} & 
            \textbf{Bronze} & 
            \textbf{Silver} & 
            \textbf{Gold} &
            \textbf{Any Medal} \\
            \midrule  
            \hspace{0.5em}\textbf{MLAB} \cite{huang2023mlagentbench} \\ 
            \hspace{0.5em}gpt-4o-2024-08-06 & 44.3 ± 2.6 & 1.9 ± 0.7 & 0.0 ± 0.0 & 0.0 ± 0.0 & 0.8 ± 0.5 & 1.3 ± 0.5 \hspace{0.5em} \\
            \cmidrule{1-7}  
            \hspace{0.5em}\textbf{OpenHands} \cite{wang2024openhands} \\ 
            \hspace{0.5em}gpt-4o-2024-08-06 & 52.0 ± 3.3 & 7.1 ± 1.7 & 0.4 ± 0.4 & 1.3 ± 0.8 & 2.7 ± 1.1 & 5.1 ± 1.3 \hspace{0.5em} \\
            \cmidrule{1-7}  
            \hspace{0.5em}\textbf{AIDE} \cite{jiang2025aide} \\ 
            \hspace{0.5em}o1-preview & 82.8 ± 1.1 & 29.4 ± 1.3 & 3.4 ± 0.5 & 4.1 ± 0.6 & 9.4 ± 0.8 & 16.9 ± 1.1 \hspace{0.5em} \\
            \cmidrule{1-7}
            \hspace{0.5em}\textbf{ML-Master} \cite{liu2025ml} \\ 
            \hspace{0.5em}deepseek-r1 & 93.3 ± 1.3 & 44.9 ± 1.2 & 4.4 ± 0.9 & 7.6 ± 0.4 & 17.3 ± 0.8 & 29.3 ± 0.8 \hspace{0.5em} \\
            \cmidrule{1-7}
            \hspace{0.5em}\textbf{Neo multi-agent} \\  
            \hspace{0.5em}undisclosed & 85.78 ± 4.45 & 40.00 ± 1.33 & 10.22 ± 2.22 & 10.22 ± 0.89 & 13.78 ± 3.55 & 34.22 ± 0.89 \hspace{0.5em} \\
            \cmidrule{1-7}
            \hspace{0.5em}\textbf{R\&D-Agent} \cite{yang2025rdagent} \\ 
            \hspace{0.5em}gpt-5 & 53.33 ± 0.00 & 40.44 ± 1.77 & 6.67 ± 2.67 & 12.00 ± 1.33 & 16.44 ± 1.77 & 35.11 ± 0.44 \hspace{0.5em} \\
            \cmidrule{1-7}
            \hspace{0.5em}\textbf{InternAgent} \cite{internagentteam2025internagent} \\ 
            \hspace{0.5em}deepseek-r1 & 96.44 ± 0.89 & 48.44 ± 2.23 & 7.11 ± 3.11 & 10.67 ± 1.34 & 18.67 ± 1.34 & 36.44 ± 1.18 \hspace{0.5em} \\
            \cmidrule{1-7}
            \hspace{0.5em}\textbf{Operand ensemble} \\ 
            \hspace{0.5em}gpt-5 (low verbosity/effort)† & 55.11 ± 14.22 & 40.89 ± 3.11 & \textbf{20.89 ± 2.22} & 7.11 ± 2.22 & 11.56 ± 0.89 & 39.56 ± 3.26 \hspace{0.5em} \\
            \cmidrule{1-7}
            \hspace{0.5em}\textbf{FM Agent} \\ 
            \hspace{0.5em}Gemini-2.5-pro & \textbf{96.89 ± 2.22} & \textbf{51.56 ± 2.23} & 8.44 ± 0.89 & \textbf{12.44 ± 3.56} & \textbf{22.67 ± 1.34} & \textbf{43.56 ± 1.78} \hspace{0.5em} \\  
            \bottomrule  
        \end{tabular}
    }
    \begin{tablenotes}
        \small
        \item † With some light assistance from an ensemble of models including Gemini-2.5-Pro, Grok-4, and Claude 4.1 Opus, distilled by Gemini-2.5-Pro.
    \end{tablenotes}
\end{table}

MLE-Bench is structured into three complexity levels: Low (dubbed Lite, containing 22 tasks), Medium (containing 38 tasks) and High (with 15 tasks). From the data presented in Figure \ref{fig:mle}, FM Agent achieved the remarkably better results on both the Medium and High complexity tasks, indicating more effective performance in the complex scenarios often encountered in real-world production environments. Furthermore, our evaluation across the complete set of 75 competitions reveals that FM Agent achieves top performance in 33 of them, surpassing all other participants on the leaderboard. For detailed results, please refer to the Appendix.

\subsection{ALE-Bench}
ALE-Bench \cite{imajuku2025ale} is designed for objective-driven algorithmic tasks, composed of computationally intractable optimization problems derived from the AtCoder Heuristic Contests, which lack known exact solutions. This format necessitates iterative solution refinement over extended time horizons to achieve higher scores, making it a suitable environment for evaluating advanced agent architectures that extend beyond simple code generation. The experimental protocol utilized the lite version of ALE-Bench, a curated subset of 10 diverse and challenging problems.

FM agent's performance was benchmarked against two baseline methods presented in the original ALE-Bench publication: an iterative refinement baseline employing a Self-Refine methodology, and the purpose-built ALE-Agent, which incorporates domain-specific knowledge and a diversity-oriented search algorithm \cite{imajuku2025ale}. To ensure a controlled comparison, all evaluated systems, including baselines and FM Agent, were configured to use Gemini 2.5 Pro as their foundational large language model.

The empirical results, summarized in Table~\ref{tab:ale}, demonstrate that:

\begin{itemize} 
    \item \textbf{State-of-the-Art Overall Performance:} FM Agent establishes a new state-of-the-art performance, achieving a mean overall score of \textbf{1976.3}. This surpasses the specialized ALE-Agent (1879.3) by 5.2\% and significantly outperforms the iterative refinement baseline (1201.3) by 64.6\%.
    \item \textbf{Superior Consistency at Expert Tiers:} While all agents achieved baseline competency ($\ge 400$ performance) on 100\% of tasks, FM Agent showed superior reliability at higher performance levels. It surpassed the $\ge 1600$ threshold on \textbf{80.0\%} of problems, compared to 70.0\% for ALE-Agent. More notably, FM Agent reached the expert "Yellow" tier ($\ge 2000$) on \textbf{50.0\%} of tasks—a substantial improvement over ALE-Agent's 30.0\%.
    \item \textbf{Exceptional Performance in Long-Horizon Tasks:} The breakdown by contest format highlights a key strength. FM Agent holds a commanding lead in long contests with an average performance of \textbf{1701.8}, compared to ALE-Agent's 1473.8. As success in these long-horizon contests often requires more sophisticated and novel solutions, this suggests FM Agent's evolutionary approach is particularly effective for the deep reasoning and refinement these complex problems demand.

\end{itemize}

\begin{table}[h]
    \centering
    \caption{FM Agent's performance on the ALE-Bench lite subset compared against the Iterative-Refinement and ALE-Agent baselines. The table details the average performance across contest formats and the distribution of performance scores across expert tiers. All agents utilize Gemini 2.5 Pro for a controlled comparison.}
    \label{tab:ale}
    \resizebox{\linewidth}{!}{
        \begin{tabular}{lccccccc}
            \toprule
            \multirow{2}{*}{\textbf{Agent}} & \multicolumn{3}{c}{\textbf{Average Perf.}} & \multicolumn{4}{c}{\textbf{Perf. Distribution (\%)}} \\
            \cmidrule(lr){2-4} \cmidrule(lr){5-8}
            & short & long & overall & $\ge 400$ & $\ge 1600$ & $\ge 2000$ & $\ge 2400$ \\
            \midrule
            Iterative-Refinement(Gemini-2.5-pro) & 1159.8 & 1242.8 & 1201.3 & \textbf{100.0} & 0.0 & 0.0 & 0.0 \\
            ALE-Agent(Gemini-2.5-pro) & \textbf{2284.8} & 1473.8 & 1879.3 & \textbf{100.0} & 70.0 & 30.0 & \textbf{20.0} \\
            FM Agent(Gemini-2.5-pro) & 2250.8 & \textbf{1701.8} & \textbf{1976.3} & \textbf{100.0} & \textbf{80.0} & \textbf{50.0} & \textbf{20.0} \\
            \bottomrule
        \end{tabular}
    }
\end{table}


\subsection{KernelBench}

\begin{table}[!t]
    \caption{Environment Setup for KernelBench.}
    \label{tab:kern_env}
    \centering
    \begin{tabular}{@{}l >{\raggedright\arraybackslash}p{0.67\textwidth}@{}}
        \toprule
        \textbf{Component} & \textbf{Specification} \\ \midrule
        CPU & Intel(R) Xeon(R) Gold 6271C CPU @ 2.60GHz \\
        GPU & NVIDIA A100 80GB \\
        OS & Ubuntu 22.04.5 LTS \\
        Packages & Python 3.12.11, PyTorch 2.6.0 with CUDA 12.4 \\
        Toolkit & CUDA Toolkit 12.4, Nsight Compute 2025.2.1, Compute Sanitizer 2023.2.0, Ninja 1.13.0 \\
        \bottomrule
    \end{tabular}
\end{table}

KernelBench \cite{kernelbench} is designed to assess LLMs’ ability to generate efficient GPU kernels. We evaluate our approach on Level 3 of KernelBench, denoting the most challenging kernels, and cluster the problems and construct a workload set that balances coverage and diversity across task types, enabling assessment of both kernel generation and end-to-end model optimization.

To better reflect the demands of real-world production, we tighten the numerical tolerance from $10^{-2}$ to $10^{-4}$, which aligns with the FP16 machine epsilon, reduces rounding errors, and maintains support for mixed precision, thus imposing stricter accuracy requirements. We also increase the warmup iterations (1 $\rightarrow$ 32) and profiling iterations (10 $\rightarrow$ 128). The benchmark environment is detailed in Table \ref{tab:kern_env}.

We compare against OpenEvolve\footnote{\url{https://github.com/codelion/openevolve}}, AI CUDA Engineer \cite{ai-cuda-engineer}, and CUDA-L1 \cite{cudal1}, representing the major lines of progress in CUDA kernel generation—agentic optimization and reinforcement learning respectively.
Specifically, AI CUDA Engineer serves as the agentic SOTA, while CUDA-L1 represents the RL-based SOTA.
Both SOTA archives \footnote{\url{https://huggingface.co/datasets/SakanaAI/AI-CUDA-Engineer-Archive}}\footnote{\url{https://github.com/deepreinforce-ai/CUDA-L1}} include solutions produced by multiple LLMs (e.g., DeepSeek v3, o3-mini, GPT-4o-mini). To ensure a strictly fair comparison, both our agent and the OpenEvolve baseline are powered by the same LLM (\texttt{Gemini-2.5-Pro}), and we execute all SOTA kernels on an A100 without modification. We further verify that, except for the kernel launcher, all generated kernels are implemented without any dependence on the \texttt{ATen} interface.

Our approach achieves \textbf{2.08$\times$ to 20.77$\times$} speedups over \texttt{torch.compile}, consistently outperforming the previous SOTA reached while maintaining numerical accuracy within $10^{-5}$.

\begin{figure*}[t]
  \centering
  \includegraphics[width=0.95\textwidth]{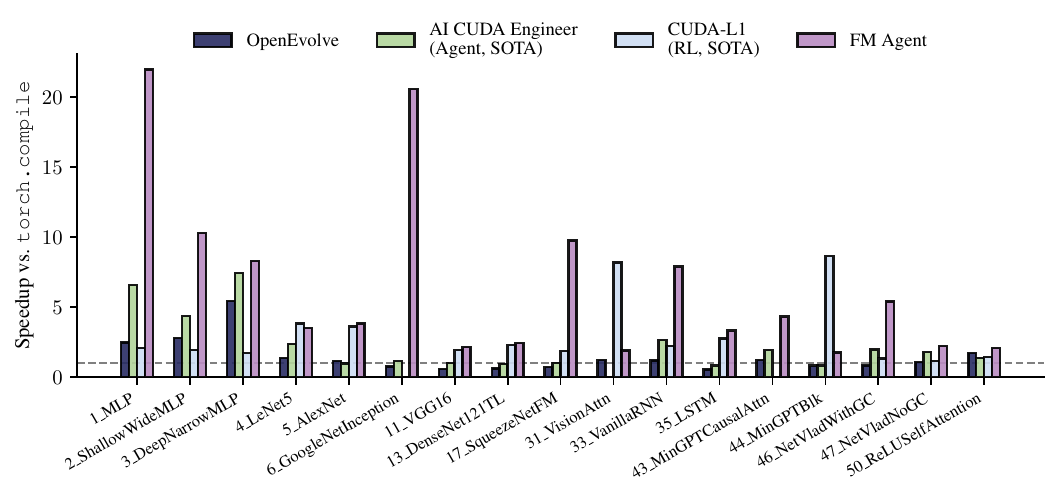}
  \caption{Comparison of speedup achieved relative to \texttt{torch.compile}. The dashed line at 1 indicates parity with \texttt{torch.compile}.}
  \label{fig:kernel}
\end{figure*}

\section{Case Study}


\subsection{Machine Learning}
\label{subsection:ml}
In machine learning, feature engineering plays a crucial role in real-world applications by transforming raw input data into informative and interpretable representations, therefore, we focus our case study on this stage. However, manual design is slow, expertise-heavy, and hard to scale; existing Auto-FE methods rely on fixed search spaces and lack domain knowledge. Recently, LLM-based attempts help with feature generation and selection~\cite{caffe,featllm,ijcai2025p314}, but reliably discovering high-value features directly from raw data remains difficult.

We introduce domain-specific evaluators and assess our framework on the American Express – Default Prediction task~\cite{amex-default-prediction}, where each sample is an \~18-month customer sequence with a default label within 120 days of the last statement. The task maps irregular sequences to fixed-size vectors via temporal transformations. To evaluate the value of feature engineering, throughout the evaluation process, the downstream model and all hyperparameters remain fixed to ensure that performance gains are solely attributable to feature improvements. 

\begin{figure}
    \centering
    \includegraphics[width=0.65\linewidth]{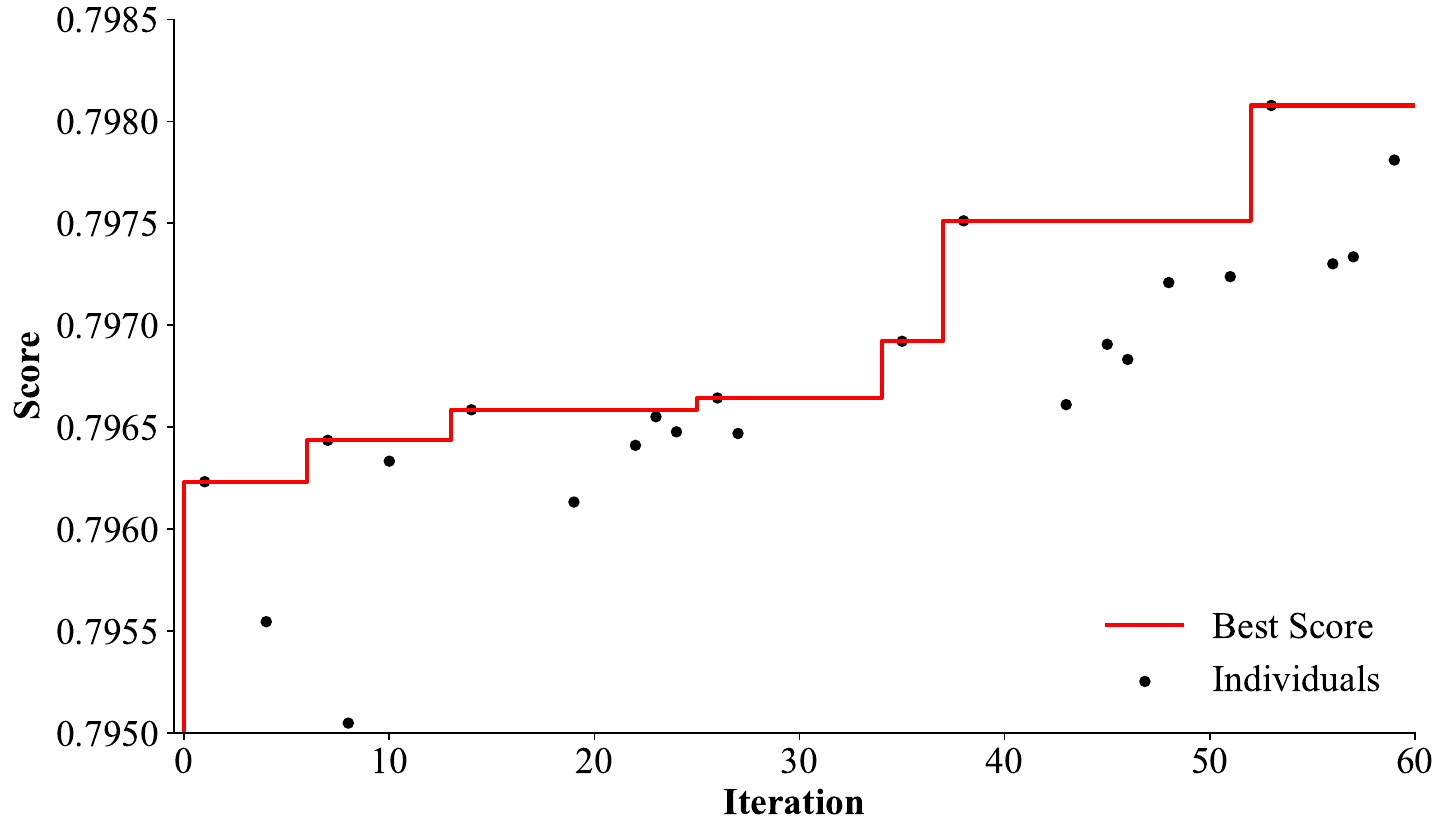}
    \caption{Convergence of the evaluation score on the American Express – Default Prediction task, where the score is the task’s original metric and higher values indicate better performance. The individual points represent the performance scores of mutated solutions in each iteration, while
the solid red line tracks the best score achieved.}
    \label{fig:case_fe}
\end{figure}

As shown in Figure~\ref{fig:case_fe}, the feature optimization process exhibits continuous performance improvement across iterations. To clarify the optimization trajectory and reporting convention, features are constructed cumulatively: each stage adds newly generated feature families to the existing set, and the reported feature counts represent post-accumulation totals. Below we summarize the cumulative feature sets produced by the agent:

\begin{itemize}
    \item \textbf{Feature Set 1 }(iteration 0, Score: 0.7951, 1,097 features)\textbf{.} Baseline features comprise numeric aggregations (mean, standard deviation, minimum, maximum, last value), categorical aggregations (count, last, number of unique values), and differentials computed as the difference between the last and penultimate record for each numeric column. These operations summarize per-customer histories into compact temporal profiles.
    \item \textbf{Feature Set 2} (iteration 7, Score: 0.7964, 1,274 features). For each numeric column, the agent computes the least-squares regression slope over the entire sequence window in a fully vectorized manner, capturing global trends across the customer’s history.
    \item \textbf{Feature Set 3} (iteration 14, Score: 0.7966, 1,474 features). Two slope-based families are added: full\_slope\_norm (slopes over the full standardized history) and recent\_slope\_norm (slopes over the most recent six periods on standardized sequences), jointly encoding long- and short-term dynamics.
    \item \textbf{Feature Set 4} (iteration 35, Score: 0.7969, 1,574 features). Numeric variables receive near-window linear trends (local polyfit slopes) and mean absolute changes, while categorical variables receive consistency (proportion of records equal to the final value) and dominance (proportion of the most frequent value), describing local temporal stability and categorical persistence.
    \item \textbf{Feature Set 5} (iteration 53, Score: 0.7981, 1,734 features). Numeric features are enriched with Exponentially Weighted Moving Averages (EWMA) and Exponentially Weighted Volatility (EW variance), emphasizing recent behavior. Categorical features include time-decayed risk mappings (using the last-record target rate per category) and weighted change counts that quantify risk-aware categorical transitions over time.

\end{itemize}

Overall, these cumulative improvements generated by FM Agent increase the end-to-end score by \textbf{+0.003}. Notably, this progress is achieved under a feature-only optimization setting, demonstrating the effectiveness and potential of our framework in the feature engineering stage of machine learning.


\subsection{Kernel Generation}
To illustrate the power of this approach, we demonstrate a live optimization session on the Flow Matching Decoder in CosyVoice2-0.5B \cite{cosyvoice2}, a critical component whose performance directly impacts user experience. Within its \texttt{flow.decoder.estimator} module, our agent automatically optimizes three representative kernels, each posing unique challenges:

\begin{itemize}
    \item \texttt{FeedForward}: FM Agent rapidly identifies operator fusion opportunities, eliminating intermediate memory traffic. Significant speedups emerge within only a few iterations, highlighting efficiency in handling common optimization patterns.

    \item \texttt{SinusoidalPosEmb}: For this lightweight scalar computation, FM Agent explores loop unrolling strategies, converging quickly on an unroll factor that maximizes instruction-level parallelism.

    \item \texttt{TimestepEmbedding}: This GEMM (General Matrix Multiplication) kernel presents a far more complex challenge. Here, FM Agent systematically explores tiling strategies and shared memory allocations. Unlike the other two kernels, improvements are slower but steady, showcasing persistence in navigating vast search spaces. Leveraging memory from past GEMM optimizations, the agent accelerates progress by transferring prior knowledge.
\end{itemize}

\begin{figure}
    \centering
    \includegraphics[width=0.75\linewidth]{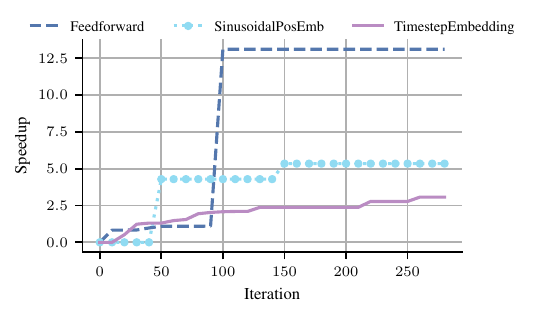}
    \caption{Speedup convergence of kernels in the CosyVoice2-0.5B Flow Matching Decoder against the official PyTorch-based implementation. FeedForward (fusion) and SinusoidalPosEmb (unrolling) converge quickly, while TimestepEmbedding shows slower convergence as it requires exploration of shared memory tiling.}
    \label{fig:case_cosy}
\end{figure}

As Figure~\ref{fig:case_cosy} illustrates, FM Agent adapts its optimization strategy to each kernel’s demands—achieving quick wins in straightforward cases while sustaining long-term exploration in complex, memory-intensive scenarios.

\subsection{Math}\label{subsection:math}

In the following, we further display the results of applying our system to three mathematical problems introduced in AlphaEvolve \cite{novikov2025alphaevolve}, where our system outperforms AlphaEvolve and achieves state-of-the-art results. As we mainly focus on complex problems in real-world R\&D processes, these early experiments on mathematical tasks primarily serve to deepen our understanding of and bring insights for the design of more effective system for practical optimization challenges encountered in real-world applications.

\renewcommand{\arraystretch}{1.3}
\begin{table}[htbp]
    \centering
    \caption{Summary of the performance metrics for three mathematical problems. The table reports the previously best-known results (prior to AlphaEvolve), the results achieved by AlphaEvolve \cite{novikov2025alphaevolve}, and our FM agent.}
    \label{tab:math-human-as-fm}
    \resizebox{\linewidth}{!}{
        \begin{tabular}{c|ccc}
            \hline
            & Previously best known & AlphaEvolve & FM \\ \hline
            Circle packing$^\dagger$ & $2.634$ & $2.635\mathbf{8627564}$  & $2.635\mathbf{9740012}$ \\
            \hline
            Ratio minimization$^*$ & $12.890$ & $12.8892\mathbf{66112}$ & $12.8892\mathbf{30201}$ \\
            \hline
            An uncertainty inequality$^*$ & $0.3523$  & $0.3520991044\textbf{225273}$ & $0.3520991044\textbf{160562}$ \\ 
            \hline
        \end{tabular}
    }
    \vspace{0.2cm}
    \begin{flushleft}
        \footnotesize
        $^\dagger$Maximization task; $^*$Minimization task.
    \end{flushleft}
\end{table}

\subsubsection{Circle Packing}\label{subsubsection:circle_packing}

\begin{figure}[htbp]
    \centering
    \begin{subfigure}[b]{0.48\columnwidth}
        \centering
        \includegraphics[width=\linewidth]{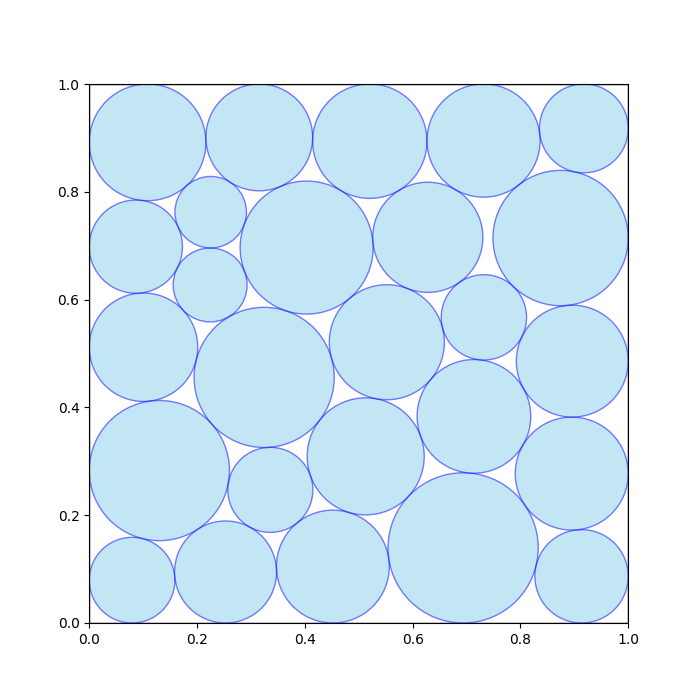}
        \caption{Final Solution for the 26-Circle Packing Problem}
        \label{subfig:circle_vis}
    \end{subfigure}
    \hfill
    \begin{subfigure}[b]{0.48\columnwidth}
        \centering
        \includegraphics[width=\linewidth]{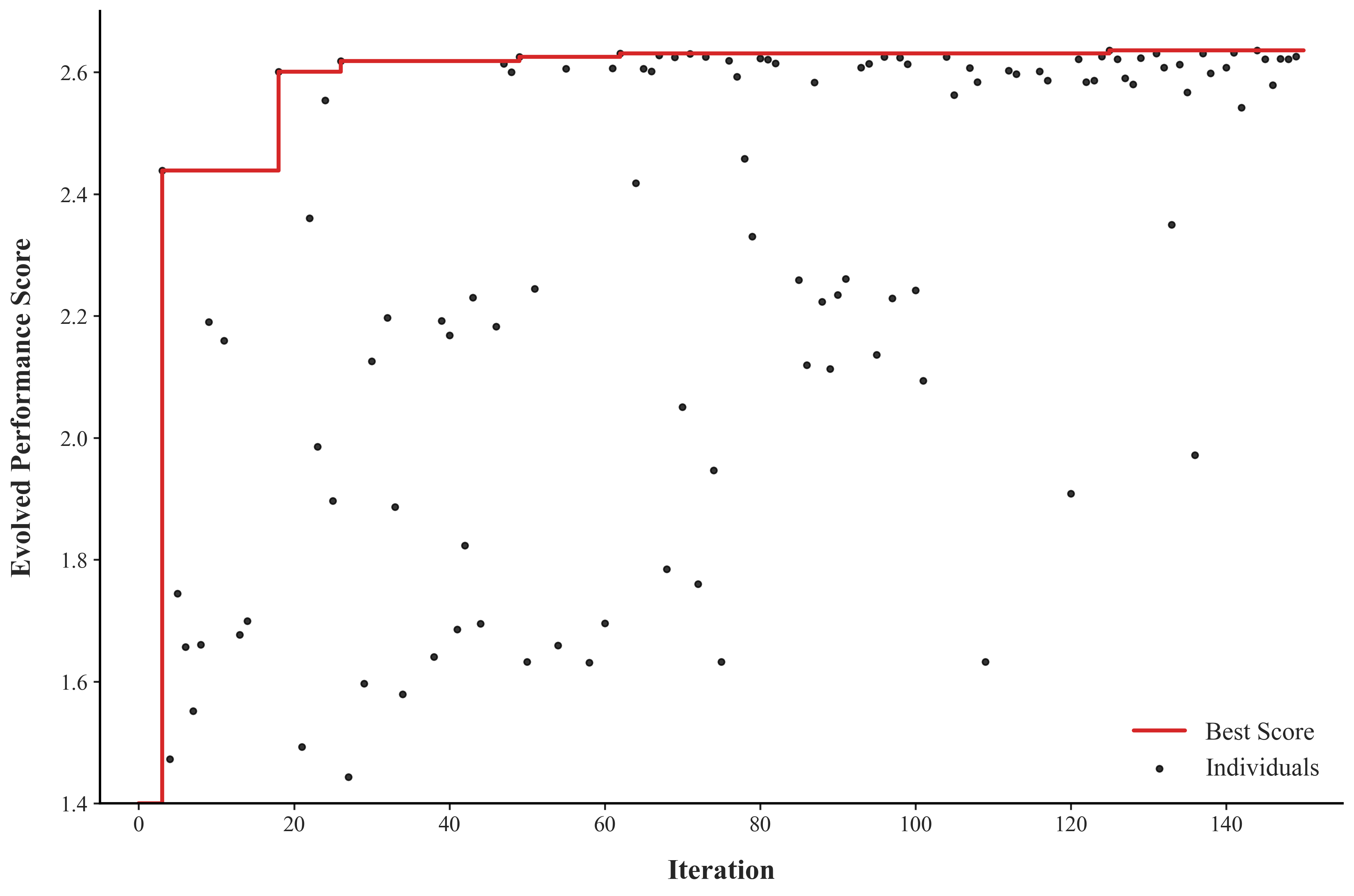}
        \caption{Evolutionary Performance on the Circle Packing Problem}
        \label{subfig:circle_evolve}
    \end{subfigure}
    \caption{Solution and Performance for the Circle Packing Task. Fig \ref{subfig:circle_vis} shows the final high-density packing configuration of 26 circles within a unit square, representing the state-of-the-art solution discovered by the FM Agent. Fig \ref{subfig:circle_evolve} illustrates the convergence of the evolutionary search over time. The individual points represent the performance scores of mutated solutions in each iteration, while the solid red line tracks the best score achieved so far by FM Agent.}
    \label{fig:circle_packing}
\end{figure}

\textbf{TASK DESCRIPTION} The circle packing optimization problem involves arranging 26 circles inside a unit square with the objective of maximizing the sum of their radii, under the constraints that no circles overlap and all remain entirely within the square's boundaries. This constrained optimization challenge integrates discrete placement choices with continuous radius adjustments, establishing it as a sophisticated benchmark in the field of evolutionary algorithms. The problem is characterized by multiple local optima, necessitating advanced search strategies to identify high-quality solutions. Naive methods tend to converge prematurely to suboptimal configurations that poorly utilize the available space.

\textbf{SOLUTION} The evolved solution constitutes a three-stage hybrid optimization methodology that yields a high-density packing configuration. The efficacy of the algorithm is rooted in its structured, sequential approach. The process commences with a geometric initialization phase, wherein a staggered, hexagonal-like lattice structure is generated to establish an advantageous initial condition consistent with dense packing theory. Subsequently, a staged, physics-informed simulation is executed, employing a two-stage gradient-based optimization routine. This routine initially applies a repulsive force to facilitate global exploration and mitigate premature convergence, followed by the introduction of an aggressive overlap penalty to refine the circle center coordinates. In the terminal stage, a Linear Programming (LP) polish is performed; for the determined set of centers, an LP solver is utilized to compute the mathematically exact optimal radii, thereby maximizing space utilization for the final configuration. 

\textbf{INSIGHTS} The evolutionary process demonstrates both the efficiency and effectiveness of FM Agent. As shown in \ref{subfig:circle_evolve}, the system converges on a near-optimal solution remarkably quickly, with the best score making significant leaps within the first 25 iterations. This rapid improvement highlights the system's efficiency in identifying promising regions of the vast search space without wasteful exploration. The performance of individual mutations, shown as scattered points, further illuminates the agent's strategy. Initially, the scores exhibit high variance, reflecting a broad exploration phase where diverse algorithmic approaches are tested. As the evolution progresses, the variance of these individual scores gradually decreases, and the points cluster more tightly around the best-known score. This transition from exploration to exploitation indicates the effectiveness of the framework's sampling and database management. The system successfully identifies high-quality parent solutions and refines them, leading to consistent, incremental improvements in the later stages. This dynamic adjustment allows the agent to navigate the complex trade-offs of the problem, ultimately discovering and refining a state-of-the-art solution as depicted in \ref{subfig:circle_vis}.

\subsubsection{An uncertainty inequality}\label{subsubsection:b4}
\textbf{TASK DESCRIPTION} 
The goal of this task is to find the minimal upper bound for a theoretically existing constant in an uncertainty inequality. Specifically, given a function $f: \mathbb{R} \to \mathbb{R}$, denote the Fourier transform $\hat{f}(\xi) := \int_{\mathbb{R}} f(x) e^{- 2 \pi i x \xi} \mathrm{d}x$ and  
\begin{displaymath}
    A(f) := \inf \{r > 0 : f (x) \ge  0  ~\text{for all}~ |x| \ge r\}.
\end{displaymath}
Denote $C^{*}$ as the largest constant such that 
\begin{displaymath}
    A(f) A(\hat{f}) \ge C^{*}, ~\text{for all even function}~ f ~\text{with}~ \max(f(0), \hat{f}(0)) < 0. 
\end{displaymath}
It is known that $0.2025 \le C^{*} \le 0.3523$ \cite{gonccalves2017hermite}. Our goal is to construct a function $f^{*}$ such that the corresponding value of $ A(f^{*}) A(\widehat{f^{*}}) $, which serves as an upper bound of $C^{*}$, is as small as possible.

AlphaEvolve \cite{novikov2025alphaevolve} further improved the upper bound to $C^{*} \le 0.3520991044225273$, while FM Agent has achieved a better solution to $ C^{*} \le 0.3520991044160562$.

\textbf{SOLUTION} 
Searching for this function over the entire function space is infeasible. Therefore, \cite{gonccalves2017hermite} proposed a construction that restricts 
the search space to functions of the form $f^{*}(x) = P(x) e^{-\pi x^{2}}$, where $P(x) = \sum_{k=0}^{K} c_{k} H_{4k}(x)$, and $H_{4k}(x)$ are Hermite polynomials.  Building on this approach, AlphaEvolve  further imposed $P(0) = 0$, set $K = 3$, and searched for the three coefficients $c_{0}, c_{1}, c_{2}$ (with $c_{3}$ fully determined by them), thereby solve the problem.

Using the same construction by \cite{gonccalves2017hermite}, our FM agent found three coefficients :
$ c_{0} \approx 4.40581122518366186113780713640$, 
$c_{1} \approx -0.1550236238960183143831272900$, 
$c_{2} \approx -0.0011938260171886596119894541$, 
thus improved the upper bound to $ C^{*} \le 0.3520991044160562$. 

Figure \ref{subfig:b4} visualizes the corresponding function. The strategy that produced the above results can be regarded as a variant of differential evolution algorithm (see Listing \ref{listing:b4-best-program} in the appendix). A closer examination of the FM Agent's evolutionary process reveals that before reaching the optimal solution, it explored several other search strategies, including random search, simulated annealing, L-BFGS (Limited-memory Broyden–Fletcher–Goldfarb–Shanno) and SLSQP (Sequential Least Squares Programming).

\textbf{INSIGHTS} 
In this task, ``An uncertainty inequality'' task involves the deliberate incorporation of domain knowledge, namely, the construction method proposed in \cite{gonccalves2017hermite}. As noted in \cite{novikov2025alphaevolve} Appendix B.4, employing more advanced constructions \cite{cohn2019optimal}, could enable AlphaEvolve to further improve the above results. Our experiment also proves that different problem settings could result in different achievements. This confirms that the search performance is highly sensitive to the problem abstractions derived from expert domain knowledge, and that different human-provided constructions can significantly impact the final results.

\subsubsection{Minimizing the ratio of maximum to minimum distance in the 2-dimensional space}\label{subsubsection:b8}

\textbf{TASK DESCRIPTION} The goal of this task is to find 16 points in the 2-dimensional space such that the ratio of the maximum and minimum pairwise distances is minimized. 

\begin{figure}[htbp]
    \centering
    \begin{subfigure}[b]{0.4\columnwidth}
        \centering
        \includegraphics[width=\linewidth]{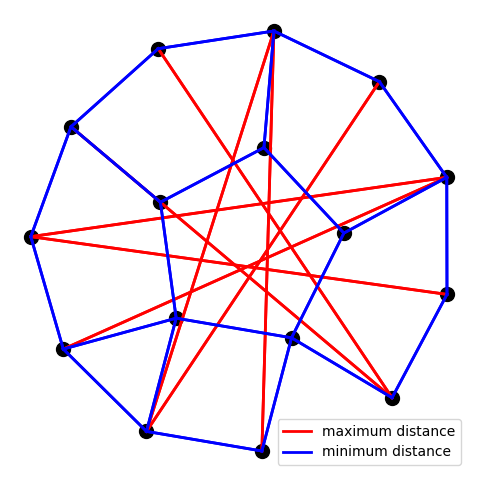}
        \caption{}
        \label{subfig:b8}
    \end{subfigure}
    \hfill
    \begin{subfigure}[b]{0.5\columnwidth}
        \centering
        \includegraphics[width=\linewidth]{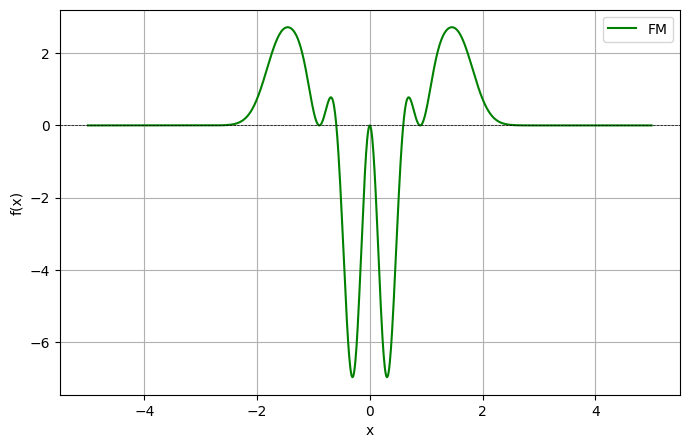}
        \caption{}
        \label{subfig:b4}
    \end{subfigure}
    \caption{Visualization of the results using the verification code provided in AlphaEvolve \cite{novikov2025alphaevolve}. Fig \ref{subfig:b8}: Minimizing the ratio of maximum to minimum distance in the 2-dimensional space; Fig \ref{subfig:b4}: An uncertainty inequality.}
    \label{fig:math-b8-b4}
\end{figure}

AlphaEvolve \cite{novikov2025alphaevolve} found 16 points with ratio $\approx \sqrt{12.889266112}$. FM Agent achieves the ratio $\approx \sqrt{12.889230201}$, with the 16 points are shown in Listing \ref{listing:b8-16-points} and visualized in Figure \ref{subfig:b8}.

\begin{figure}[htbp]
\centering
\begin{lstlisting}[basicstyle=\ttfamily\footnotesize,language={}, caption={16 points in 2D space founded by FM agent.}, label={listing:b8-16-points}]
[-1.47975561, 0.98098357], [ 0.85184808, 0.07211039], [-0.73461161, 1.64788717], 
[ 0.15127319,-1.78975576], [-0.71559290, 0.33596005], [-1.54547711,-0.91892085], 
[ 1.73300231,-0.45160218], [-1.82309280, 0.04177136], [ 1.73113866, 0.54839609], 
[ 1.15533190, 1.36598190], [ 0.40491725,-0.82245813], [-0.58095076,-0.65493425],
[ 0.16887753, 0.80255630], [ 0.25391499, 1.79893406], [-0.83459483,-1.62223187], 
[ 1.26377170,-1.33467785]
\end{lstlisting}
\end{figure}

\textbf{SOLUTION} 
Delving into the specific strategy for generating these 16 points (see Listing \ref{listing:b8-best-program} in the appendix), the optimization is 
performed using the SLSQP method (Sequential Least Squares Programming), 
a gradient-based algorithm suitable for smooth constrained optimization problems. To enhance robustness and reduce the risk of convergence to poor local minima, a multi-start strategy is employed with five diverse initial configurations: a hexagonal lattice, a 1–5–10 concentric ring structure, a 6–10 two-ring structure, a 4$\times$4 regular grid, and a random layout. Each configuration is independently optimized, and the one yielding the lowest maximum-to-minimum distance ratio is selected as the final solution. Additionally, the program employs a vectorized Jacobian for the constraint gradients, which improves computational efficiency when evaluating the many pairwise distance constraints.

\textbf{INSIGHTS}
In this task, we find that even without formal proofs, high-level guidance inspired by human intuition can effectively constrain the search space by informing the system setup. By seeding the optimization with several diverse structures, such as a hexagonal lattice, concentric rings, regular grids, and random layouts, FM Agent can explore promising regions more effectively and avoid poor local minima. Compared with fully zero initialization, these coarse, expert-inspired hints reduce the effective search space while still allowing the agent to refine the solution, achieving similarly strong or better performance. This demonstrates that even partial domain knowledge, such as rough spatial arrangements, can substantially guide the agent and improve overall optimization results.


\section{Ablations}
\begin{figure}[t]
    \centering
    \begin{subfigure}[t]{0.6\textwidth} 
        \centering
        \includegraphics[width=\textwidth]{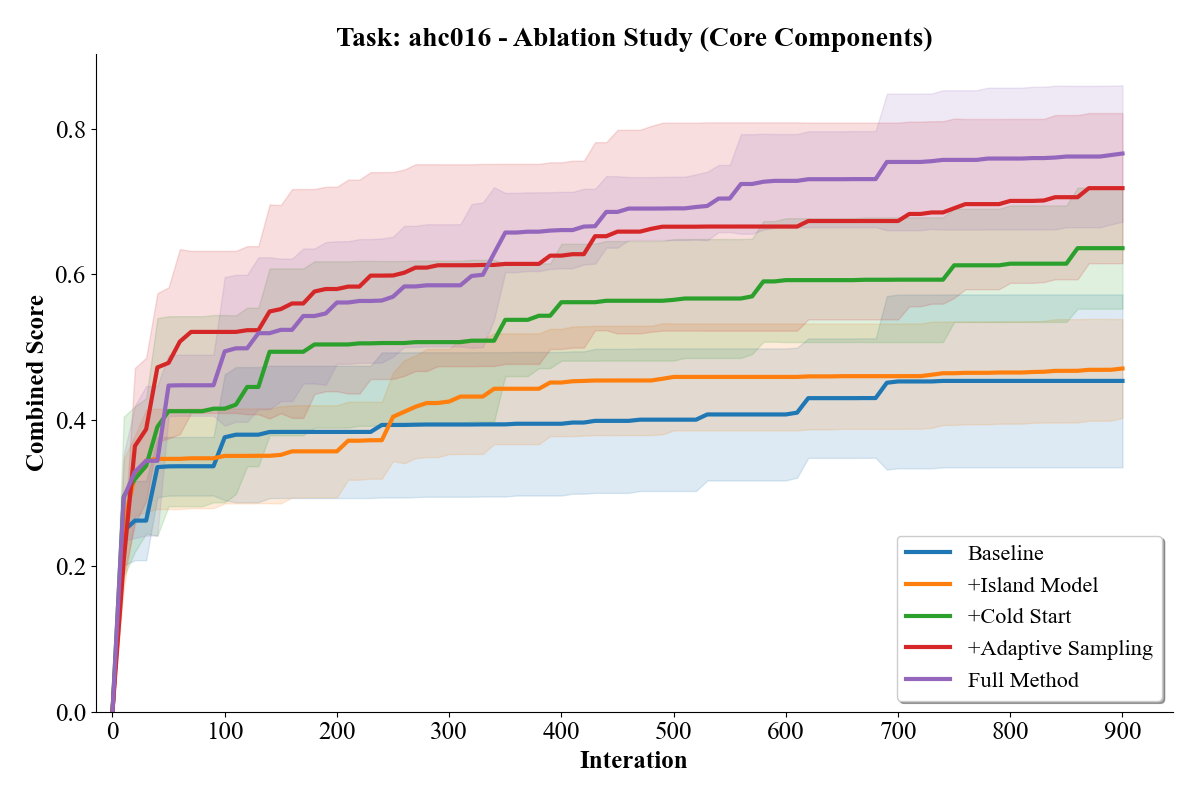}
        \label{fig:subfig:a}
    \end{subfigure}
    \caption{Ablation study results of FM Agent on the ahc016 task. Each curve displays the performance of a different experimental setting, averaged over five independent runs (where a higher combined score is better). The shading indicates the standard deviation.} 
    \label{fig:combined_ablation}
\end{figure}

\begin{figure}[t]
    \centering
    \begin{subfigure}[t]{0.49\textwidth} 
        \centering
        \includegraphics[width=\textwidth]{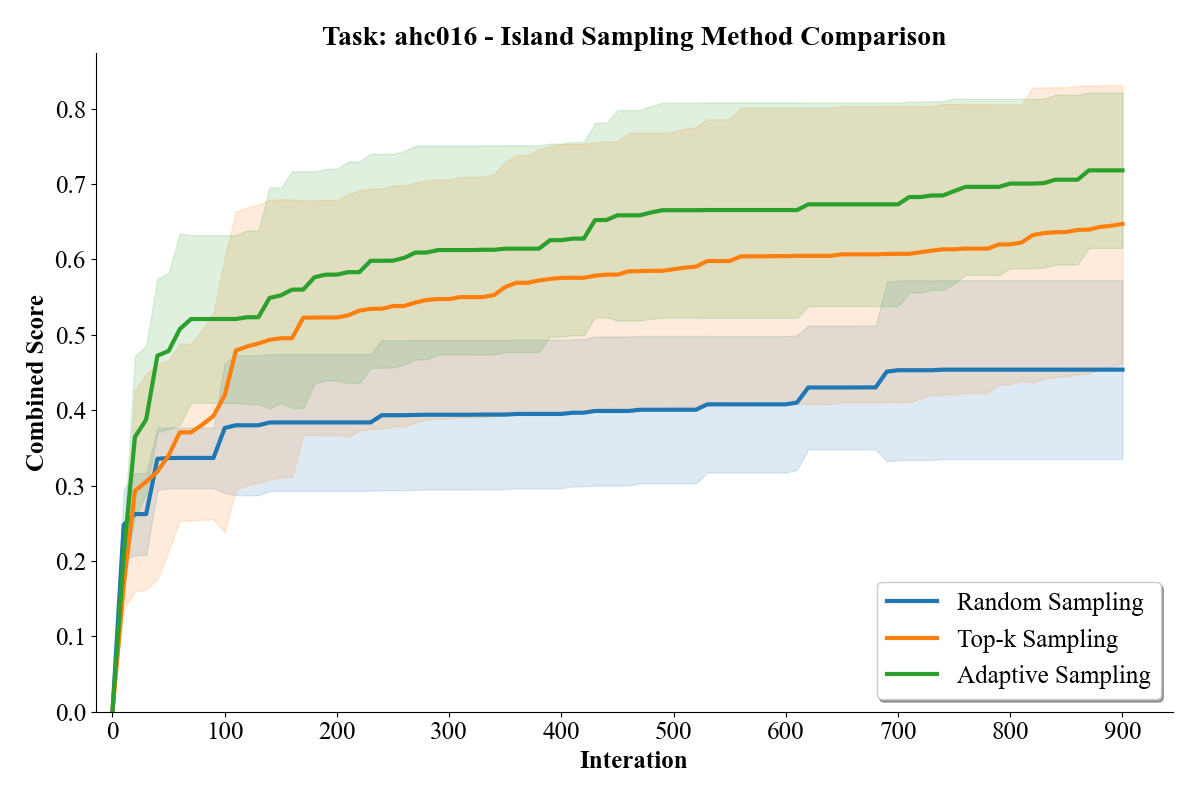}
        \label{fig:subfig:a}
    \end{subfigure}
    \hfill 
    \begin{subfigure}[t]{0.49\textwidth} 
        \centering
        \includegraphics[width=\textwidth]{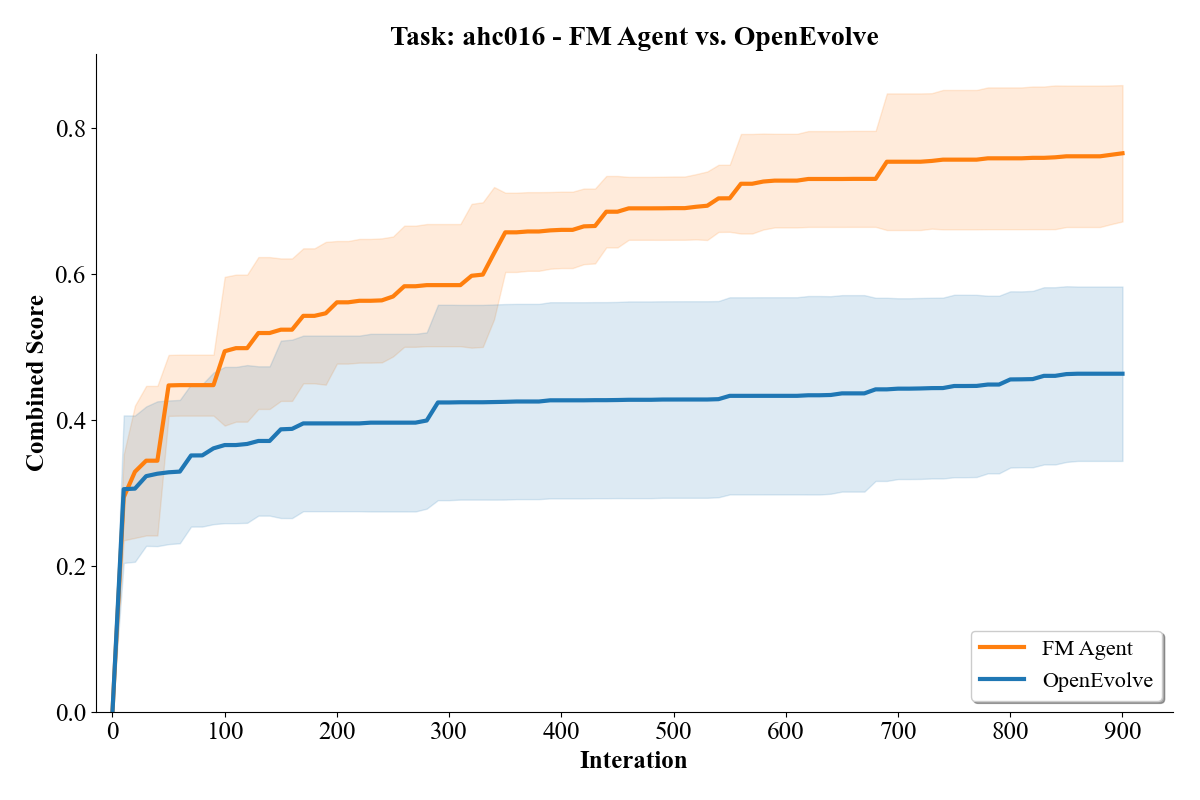}
        \label{fig:subfig:b}
    \end{subfigure}
    \caption{\textbf{Left}: Comparison of different sampling methods on the ahc016 task. \textbf{Right}: Compared with open-source baseline on the ahc016 task.  Each curve displays the performance of a different experimental setting, averaged over five independent runs (where a higher combined score is better). The shading indicates the standard deviation.}
    \label{fig:ablation}
\end{figure}

To rigorously quantify the contributions of our proposed components, we conducted a comprehensive ablation study against the baseline system. All evaluations were performed on a selected task, ahc016 \footnote{\url{https://atcoder.jp/contests/ahc016/tasks/ahc016_a}}, from ALE-Bench. The competition requires encoding integers as graphs so that, after noise and vertex shuffling, the original integer can be accurately recovered. Participants must generate a set of graphs and then predict the source of each noisy graph, aiming to maximize accuracy while minimizing graph size.

Due to the inherent stochasticity in our evolutionary framework and the relatively high computational cost of each run, we executed each experimental setting five times on the same task and report the averaged results. This procedure ensures that the observed performance differences reliably reflect the contributions of individual components rather than random fluctuations.

\textbf{Overall Contribution.}
As shown in Figure \ref{fig:combined_ablation}, quantitative comparison demonstrates that each proposed component—adaptive sampling, cold start, and island model—meaningfully contributes to the overall performance improvement. The Full Method, which integrates all components, achieves the best overall performance, confirming a powerful synergistic effect by combining their individual strengths to surpass all other configurations in the high final score. Specifically, the cold start strategy provided a significant initial advantage, accelerating the system's early convergence rate. The adaptive sampling strategy continuously drives performance improvement by achieving a better balance between exploration and exploitation. The island model strategy, by promoting solution diversity, rapidly reached the baseline's peak performance level during the intermediate phase.

\textbf{Adaptive Sampling Strategy.} We benchmarked the adaptive sampling method against two baselines, including random sampling and top-k sampling, in $\text{Figure \ref{fig:ablation}}$ (Left). Random sampling method selects programs uniformly, emphasizing pure exploration, while top-k sampling method propagates only the top-$\text{k}$ programs, representing pure exploitation. Our method achieves a final combined score of $\mathbf{0.7182}$, outperforming top sampling by $\mathbf{10.99\%}$ and random sampling by $\mathbf{58.26\%}$. Notably, the adaptive sampling surpasses the maximum score of random sampling as early as $\mathbf{Iteration 40}$, demonstrating faster convergence (vs $\mathbf{Iteration 900}$ for random sampling) and a higher performance ceiling.

\textbf{Comparison with Open-Source Baseline.} The comparison of our complete FM Agent framework against OpenEvolve\footnote{\url{https://github.com/codelion/openevolve}} is presented in $\text{Figure \ref{fig:ablation}}$ (Right). The results confirm the effectiveness of our integrated design: FM Agent achieves faster convergence, higher final scores, and a more robust evolutionary trajectory.

\section{Related work}
\subsection{Multi-Agent Systems}
The complexity of modern software development has driven a shift from single-agent solutions to multi-agent collaboration paradigms, mirroring human development teams. Initial pioneering systems, such as \cite{dong2024self, huang2024agentcoder, rasheed2024codepori}, established structured, specialized pipelines with fixed roles (e.g., analysis, coding, testing). Subsequent research aimed to enhance the efficiency and reliability of these collaborations, introducing mechanisms like "dual-collaboration" \cite{chen2023gamegpt} for information reuse, declarative memory modules \cite{qi2024cleanagent} to reduce redundancy, and rigorous generate-test-fix loops (e.g., LingMa \cite{ma2024lingma}, CodeCor \cite{pan2025codecor}). Moving beyond simple code generation to complex R\&D and optimization tasks, efforts have focused on systems employing specialized collaborative roles. For instance, R\&D Agent \cite{yang2025rdagent} utilizes a dual Researcher and Developer agent collaboration, enhancing diversity by periodically merging superior results, and MLE-STAR \cite{nam2025mle} integrates collaborative agents with external knowledge, using ablation study guided refinement for targeted code improvement. More recently, systems like CoMAS \cite{xue2025comascoevolvingmultiagentsystems} explored autonomous improvement by learning from interactions, and InternAgent \cite{internagentteam2025internagent} introduced a unified closed-loop research framework that seamlessly integrates idea generation, experiment execution, and result feedback. However, a limitation of most prior work is their primary focus on code generation tasks, and FM Agent addresses this by shifting the focus from simply generating correct code to autonomously discovering state-of-the-art solutions in a large-scale searching way across diverse, complex problem domains and real-world scenarios.

\subsection{Search-driven and Evolutionary Paradigms}
The Search-driven Paradigm leverages the generative power of Large Language Models (LLMs) combined with iterative evolutionary or reinforcement-style loops to systematically explore vast solution spaces. This direction was notably established by foundational works such as FunSearch \cite{romera2024mathematical}, which introduced a generate-score-evolve closed-loop mechanism, and EoH \cite{liu2024evolution}, which proposed a "Thought-Code" dual-evolution framework. Subsequent research scaled these concepts through frameworks like AlphaEvolve \cite{novikov2025alphaevolve} for modifying entire codebases. Other efforts enhanced the search efficacy and strategy by shifting the evolutionary focus from individual solutions to the search space itself, such as X-evolve \cite{zhai2025x} which generates tunable, parameterized programs, and C-Evolve \cite{li2025c} which evolves consensus-driven prompts. Refinements to search strategy include the parallel beam-search strategy employed by ALE-Bench \cite{imajuku2025ale} to explore multiple promising paths, and ML-Master \cite{liu2025ml} which integrates Monte Carlo Tree Search (MCTS) inspired exploration with an adaptive memory mechanism. ShinkaEvolve\cite{lange2025shinkaevolve} improves sample efficiency through synergistic use of parent program sampling strategy, rejection-sampling, and LLM ensemble selection. We further optimized this paradigm by integrating expert-guided cold-start initialization to jumpstart the search, a novel adaptive diversity-driven sampling strategy for superior exploration-exploitation balance, and a distributed, asynchronous Ray infrastructure that enables the required large-scale search, which is essential for achieving state-of-the-art results on industrial-scale scenarios.

\bibliography{ref}

\newpage
\appendix
\appendix

\subsection*{A. MLE-Bench: Full results}
\textbf{Full table of results.}We present the complete scores of FM Agent across all 75 competitions in MLE-Bench. In accordance with the official MLE-Bench protocol requiring three independent submissions per competition, we report the highest score achieved across these three runs for each competition. These results are compared against several top-performing participants on the leaderboard, including Operand ensemble, InternAgent, R\&D-Agent, Neo multi-agent, and ML-Master.
As shown in the table, FM Agent achieved first place among all agents in 33 out of the 75 competitions. Its performance is even more pronounced on the hard split, where it secured first place in 9 out of the 15 total challenges. This result strongly demonstrates FM Agent’s enhanced capability in handling complex tasks.

\begin{table}[h]
    \caption{FM Agent surpasses all baseline across every evaluation dimension defined in MLE-Bench. All values represent competition-specific metrics (consistent with official MLE-Bench definitions). The results for MLAB(gpt-4o-2024-08-06), OpenHands(gpt-4o-2024-08-06), AIDE(o1-preview), R\&D-Agent(gpt-5), ML-Master(deepseek-r1), Neo multi-agent, InterAgent(deepseek-r1) and Operand ensemble(gpt-5, low verbosity/effort) are taken from the official MLE-Bench report. Results for FM Agent are averaged over three independent runs with different random seeds and are presented as the mean ± one standard error of the mean (SEM). The best-performing model in each category is highlighted in bold.(Continued on next page)}
    \label{tab:mle-leaderboard}
    \centering 
    \setlength{\heavyrulewidth}{1.5pt} 
    \setlength{\lightrulewidth}{0.5pt}   
    \resizebox{\linewidth}{!}{
        \begin{tabular}{@{}lcccccc@{}}  
            \toprule  
            \textbf{Competition} &  
            \textbf{FM Agent} &  
            \textbf{Operand ensemble} & 
            \textbf{InternAgent \cite{internagentteam2025internagent}} & 
            \textbf{R\&D-Agent \cite{yang2025rdagent}} &
            \textbf{Neo multi-agent} &
            \textbf{ML-Master \cite{liu2025ml}} \\
            \midrule  

            \textbf{aerial-cactus-identification} & $\textbf{1.0}$ & $\textbf{1.0}$ & $\textbf{1.0}$ & $\textbf{1.0}$ & $\textbf{1.0}$ & $\textbf{1.0}$ \\
            \textbf{aptos2019-blindness-detection} & $\textbf{0.93556}$ & $0.9039$ & $0.92125$ & $0.92034$ & $0.92498$ & $0.92954$ \\
            \textbf{denoising-dirty-documents$^*$} & $0.01311$ & $0.01081$ & $0.0116$ & $0.01009$ & $\textbf{0.00742}$ & $0.00763$ \\
            \textbf{detecting-insults-in-social-commentary} & $\textbf{0.95712}$ & $0.9128$ & $0.94484$ & $0.94918$ & $None$ & $0.94838$ \\
            \textbf{dog-breed-identification$^*$} & $0.3502$ & $0.53653$ & $\textbf{0.30026}$ & $None$ & $0.43795$ & $0.32345$ \\
            \textbf{dogs-vs-cats-redux-kernels-edition$^*$} & $0.00945$ & $0.02172$ & $\textbf{0.00287}$ & $0.0117$ & $0.00792$ & $0.00309$ \\
            \textbf{histopathologic-cancer-detection} & $0.99393$ & $0.93029$ & $\textbf{0.99833}$ & $0.99604$ & $0.99521$ & $0.99754$ \\
            \textbf{jigsaw-toxic-comment-classification-challenge} & $0.98694$ & $0.98047$ & $\textbf{0.98732}$ & $0.98661$ & $0.98713$ & $0.9864$ \\
            \textbf{leaf-classification$^*$} & $0.11865$ & $0.00883$ & $0.01856$ & $\textbf{0.00158}$ & $0.2459$ & $0.01661$ \\
            \textbf{mlsp-2013-birds} & $0.91413$ & $0.90827$ & $0.88582$ & $0.90731$ & $\textbf{0.93904}$ & $0.78133$ \\
            \textbf{new-york-city-taxi-fare-prediction$^*$} & $\textbf{3.96969}$ & $4.69105$ & $5.75886$ & $None$ & $None$ & $6.125$ \\
            \textbf{nomad2018-predict-transparent-conductors$^*$} & $\textbf{0.05259}$ & $0.06018$ & $0.0585$ & $0.05833$ & $0.05978$ & $0.059$ \\
            \textbf{plant-pathology-2020-fgvc7} & $\textbf{0.99957}$ & $0.98971$ & $0.99688$ & $0.99818$ & $0.99752$ & $0.99041$ \\
            \textbf{random-acts-of-pizza} & $0.7588$ & $0.70063$ & $0.67654$ & $\textbf{0.79623}$ & $0.79216$ & $0.6516$ \\
            \textbf{ranzcr-clip-catheter-line-classification} & $\textbf{0.96028}$ & $0.95833$ & $0.931$ & $None$ & $0.95585$ & $0.95979$ \\
            \textbf{siim-isic-melanoma-classification} & $0.9279$ & $0.76117$ & $\textbf{0.9414}$ & $None$ & $0.84593$ & $0.91251$ \\
            \textbf{spooky-author-identification$^*$} & $0.23578$ & $0.27869$ & $\textbf{0.21544}$ & $0.22926$ & $0.23527$ & $0.26013$ \\
            \textbf{tabular-playground-series-dec-2021} & $0.96137$ & $\textbf{0.96335}$ & $\textbf{0.96302}$ & $0.96312$ & $0.95998$ & $\textbf{0.96302}$ \\
            \textbf{tabular-playground-series-may-2022} & $\textbf{0.99566}$ & $0.69736$ & $0.9945$ & $None$ & $0.99296$ & $0.99463$ \\
            \textbf{text-normalization-challenge-english-language} & $0.99059$ & $0.99221$ & $0.99053$ & $\textbf{0.99283}$ & $0.95377$ & $0.99182$ \\
            \textbf{text-normalization-challenge-russian-language} & $0.97021$ & $0.97968$ & $0.97924$ & $0.98277$ & $\textbf{0.98304}$ & $0.97348$ \\
            \textbf{the-icml-2013-whale-challenge-right-whale-redux} & $0.99391$ & $0.94555$ & $\textbf{0.99413}$ & $0.99261$ & $None$ & $0.99082$ \\

            \textbf{AI4Code} & $0.503$ & $None$ & $0.62471$ & $None$ & $0.40432$ & $\textbf{0.72299}$ \\ 
            \textbf{alaska2-image-steganalysis} & $\textbf{0.86242}$ & $None$ & $0.62477$ & $None$ & $0.61367$ & $0.76772$ \\ 
            \textbf{billion-word-imputation$^*$} & $\textbf{6.69267}$ & $None$ & $None$ & $None$ & $7.0074$ & $None$ \\ 
            \textbf{cassava-leaf-disease-classification} & $\textbf{0.9006}$ & $0.89948$ & $0.89649$ & $0.88976$ & $0.89126$ & $0.89163$ \\ 
            \textbf{cdiscount-image-classification-challenge} & $\textbf{0.72038}$ & $None$ & $0.64694$ & $None$ & $0.65588$ & $0.65574$ \\ 
            \textbf{chaii-hindi-and-tamil-question-answering} & $\textbf{0.7433}$ & $0.7189$ & $0.27026$ & $None$ & $0.72366$ & $0.62283$ \\ 
            \textbf{champs-scalar-coupling$^*$} & $1.23566$ & $1.99777$ & $1.43002$ & $None$ & $\textbf{1.15637}$ & $1.7062$ \\ 
            \textbf{facebook-recruiting-iii-keyword-extraction} & $\textbf{0.57231}$ & $0.1991$ & $0.52234$ & $None$ & $0.53506$ & $0.4469$ \\ 
            \textbf{freesound-audio-tagging-2019} & $0.69921$ & $0.68063$ & $\textbf{0.71638}$ & $0.68569$ & $0.59279$ & $0.68586$ \\ 
            \textbf{google-quest-challenge} & $0.41448$ & $0.4105$ & $0.41797$ & $0.42004$ & $\textbf{0.43395}$ & $0.41889$ \\ 
            \textbf{h-and-m-personalized-fashion-recommendations} & $\textbf{0.02714}$ & $None$ & $0.02456$ & $0.0$ & $0.02411$ & $0.02518$ \\ 
            \textbf{herbarium-2020-fgvc7} & $\textbf{0.47074}$ & $0.09606$ & $0.38989$ & $0.45351$ & $0.44275$ & $0.23544$ \\ 
            \textbf{herbarium-2021-fgvc8} & $0.32257$ & $0.34083$ & $0.28441$ & $0.23133$ & $\textbf{0.43683}$ & $0.37574$ \\ 
            \textbf{herbarium-2022-fgvc9} & $\textbf{0.77877}$ & $0.65859$ & $0.48576$ & $0.31697$ & $0.7735$ & $0.61428$ \\ 
            \textbf{hotel-id-2021-fgvc8} & $0.44257$ & $0.22754$ & $\textbf{0.70941}$ & $0.49623$ & $0.43788$ & $0.61294$ \\ 
            \textbf{hubmap-kidney-segmentation} & $0.0$ & $None$ & $0.92562$ & $\textbf{0.9991}$ & $0.05064$ & $0.04435$ \\ 
            \textbf{icecube-neutrinos-in-deep-ice$^*$} & $1.53458$ & $None$ & $\textbf{1.51821}$ & $None$ & $None$ & $1.55861$ \\ 
            \textbf{imet-2020-fgvc7} & $0.60932$ & $0.27955$ & $\textbf{0.62573}$ & $None$ & $0.5865$ & $0.60408$ \\ 
            \textbf{inaturalist-2019-fgvc6$^*$} & $0.99814$ & $0.13767$ & $\textbf{0.13544}$ & $0.21165$ & $0.19454$ & $0.18964$ \\ 
            \textbf{iwildcam-2020-fgvc7} & $\textbf{0.83522}$ & $0.73007$ & $0.8288$ & $0.70202$ & $0.74062$ & $0.82057$ \\ 
            \textbf{jigsaw-unintended-bias-in-toxicity-classification} & $0.80032$ & $None$ & $0.78648$ & $None$ & $0.81502$ & $\textbf{0.82847}$ \\ 
            \textbf{kuzushiji-recognition} & $0.68409$ & $0.72021$ & $0.58578$ & $0.83033$ & $\textbf{0.95238}$ & $0.62457$ \\ 
            \textbf{learning-agency-lab-automated-essay-scoring-2} & $\textbf{0.84839}$ & $0.83013$ & $0.83995$ & $0.83751$ & $0.83881$ & $0.82104$ \\ 
            \textbf{lmsys-chatbot-arena$^*$} & $1.00886$ & $None$ & $1.00457$ & $None$ & $\textbf{0.99092}$ & $1.05199$ \\ 
            \textbf{multi-modal-gesture-recognition$^*$} & $0.86342$ & $\textbf{0.5383}$ & $0.6872$ & $None$ & $0.86824$ & $0.90841$ \\ 
            \textbf{osic-pulmonary-fibrosis-progression} & $-7.62814$ & $\textbf{-7.19604}$ & $-7.22947$ & $None$ & $-8.52062$ & $None$ \\ 
            \textbf{petfinder-pawpularity-score$^*$} & $17.47187$ & $\textbf{16.89225}$ & $18.19441$ & $None$ & $18.77071$ & $17.80862$ \\ 
            \textbf{plant-pathology-2021-fgvc8} & $0.9226$ & $\textbf{0.93141}$ & $0.91868$ & $0.91845$ & $0.9156$ & $0.93039$ \\ 
            \textbf{seti-breakthrough-listen} & $\textbf{0.86216}$ & $0.52149$ & $0.84358$ & $0.77904$ & $0.79164$ & $0.83485$ \\ 
            \textbf{statoil-iceberg-classifier-challenge$^*$} & $0.6963$ & $None$ & $\textbf{0.19655}$ & $None$ & $0.22671$ & $0.24349$ \\ 
            \textbf{tensorflow-speech-recognition-challenge} & $0.33647$ & $None$ & $0.3527$ & $None$ & $\textbf{0.35316}$ & $0.34837$ \\ 
            \textbf{tensorflow2-question-answering} & $\textbf{0.57823}$ & $None$ & $0.57117$ & $None$ & $0.57117$ & $0.2207$ \\ 
            \textbf{tgs-salt-identification-challenge} & $0.7127$ & $None$ & $\textbf{0.7927}$ & $None$ & $0.7831$ & $0.5221$ \\ 
            \textbf{tweet-sentiment-extraction} & $0.71906$ & $0.71958$ & $0.64334$ & $0.71159$ & $\textbf{0.75393}$ & $0.70943$ \\ 
            \textbf{us-patent-phrase-to-phrase-matching} & $\textbf{0.87466}$ & $0.84507$ & $0.87046$ & $0.84449$ & $0.83287$ & $0.85901$ \\ 
            \textbf{uw-madison-gi-tract-image-segmentation} & $0.75471$ & $None$ & $\textbf{0.83297}$ & $None$ & $0.44372$ & $0.10015$ \\ 
            \textbf{ventilator-pressure-prediction$^*$} & $\textbf{0.22428}$ & $17.65486$ & $0.47338$ & $None$ & $0.90693$ & $0.33547$ \\ 
            \textbf{whale-categorization-playground} & $\textbf{0.50367}$ & $0.43053$ & $0.31698$ & $0.27506$ & $0.42718$ & $0.2423$ \\ 
            \bottomrule  
        \end{tabular}
    }
    \vspace{0.2cm}
    \footnotesize 
    $^*$ Indicates minimization tasks, where smaller values represent better performance. \\ 
    $None$ Indicates that either no valid submission file was generated or the file format was incorrect.
\end{table}

\begin{table}[h]
\ContinuedFloat 
    \caption{FM Agent surpasses all baseline models across every evaluation dimension defined in MLE-Bench. All values represent competition-specific metrics (consistent with official MLE-Bench definitions). The results for MLAB(gpt-4o-2024-08-06), OpenHands(gpt-4o-2024-08-06), AIDE(o1-preview), R\&D-Agent(gpt-5), ML-Master(deepseek-r1), Neo multi-agent, InterAgent(deepseek-r1) and Operand ensemble(gpt-5, low verbosity/effort) are taken from the official MLE-Bench report. Results for FM Agent are averaged over three independent runs with different random seeds and are presented as the mean ± one standard error of the mean (SEM). The best-performing model in each category is highlighted in bold.}
    \label{tab:mle-leaderboard}
    \centering 
    \setlength{\heavyrulewidth}{1.5pt} 
    \setlength{\lightrulewidth}{0.5pt}   
    \resizebox{\linewidth}{!}{
        \begin{tabular}{@{}lcccccc@{}}  
            \toprule  
            \textbf{Competition} &  
            \textbf{FM Agent} &  
            \textbf{Operand ensemble} & 
            \textbf{InternAgent \cite{internagentteam2025internagent}} & 
            \textbf{R\&D-Agent \cite{yang2025rdagent}} &
            \textbf{Neo multi-agent} &
            \textbf{ML-Master \cite{liu2025ml}} \\
            \midrule  
            \textbf{3d-object-detection-for-autonomous-vehicles} & $0.0$ & $None$ & $0.0$ & $\textbf{0.01299}$ & $0.0$ & $0.0$ \\ 
            \textbf{bms-molecular-translation$^*$} & $\textbf{42.11198}$ & $None$ & $73.919$ & $None$ & $89.48078$ & $96.89238$ \\ 
            \textbf{google-research-identify-contrails-reduce-global-warming} & $0.01126$ & $None$ & $\textbf{0.07186}$ & $None$ & $0.05841$ & $0.42954$ \\ 
            \textbf{hms-harmful-brain-activity-classification$^*$} & $\textbf{0.65027}$ & $None$ & $0.75958$ & $None$ & $0.83488$ & $0.93875$ \\ 
            \textbf{iwildcam-2019-fgvc6} & $0.37797$ & $0.39804$ & $0.29297$ & $0.4711$ & $0.47775$ & $\textbf{0.48054}$ \\ 
            \textbf{nfl-player-contact-detection} & $\textbf{0.64045}$ & $None$ & $0.60575$ & $None$ & $0.0604$ & $0.54446$ \\ 
            \textbf{predict-volcanic-eruptions-ingv-oe$^*$} & $3705196.0$ & $2188576$ & $2499380.0$ & $\textbf{2044627.0}$ & $2902638.0$ & $2826639.0$ \\ 
            \textbf{rsna-2022-cervical-spine-fracture-detection$^*$} & $0.59289$ & $0.69315$ & $0.56419$ & $None$ & $\textbf{0.5639}$ & $0.5723$ \\ 
            \textbf{rsna-breast-cancer-detection} & $\textbf{0.08437}$ & $0.06863$ & $0.06232$ & $None$ & $0.04624$ & $0.04901$ \\ 
            \textbf{rsna-miccai-brain-tumor-radiogenomic-classification} & $\textbf{0.66059}$ & $None$ & $0.59647$ & $0.58588$ & $0.60941$ & $0.61176$ \\ 
            \textbf{siim-covid19-detection} & $\textbf{0.44507}$ & $0.30458$ & $0.42364$ & $None$ & $0.19073$ & $0.36954$ \\ 
            \textbf{smartphone-decimeter-2022$^*$} & $\textbf{5.9367}$ & $None$ & $6.09075$ & $None$ & $3122773.6564$ & $15642.40004$ \\ 
            \textbf{stanford-covid-vaccine$^*$} & $0.31981$ & $0.31841$ & $\textbf{0.22919}$ & $0.24187$ & $0.28491$ & $0.23945$ \\ 
            \textbf{vesuvius-challenge-ink-detection} & $\textbf{0.48622}$ & $0.20266$ & $0.19222$ & $None$ & $0.14406$ & $0.12163$ \\ 
            \textbf{vinbigdata-chest-xray-abnormalities-detection} & $\textbf{0.29848}$ & $None$ & $None$ & $None$ & $0.02371$ & $None$ \\ 
            
            \bottomrule  
        \end{tabular}
    }
    \vspace{0.2cm}
    \footnotesize 
    $^*$ Indicates minimization tasks, where smaller values represent better performance. \\ 
    $None$ Indicates that either no valid submission file was generated or the file format was incorrect.
\end{table}


\subsection*{B. Feature Mining: American Express Default Prediction task}
As discussed in Section~\ref{subsection:ml}, we illustrate FM Agent’s automated feature construction using Feature Set~5.
Each sample is a customer’s irregular sequence of statements ordered by the statement date $S_2$ (the per-record timestamp). For a given numerical time series $\{x_1,x_2,\ldots,x_n\}$, where $x_i$ denotes the value at the $i$-th record in chronological order, $i\in\{1,\ldots,n\}$, and $n$ is the number of records for that customer, the agent applies an exponential recency scheme governed by a decay hyperparameter $\alpha=0.3$. The temporal weight $w_i$ for record $i$ is
\[
w_i \;=\; (1-\alpha)^{(n-i)} ,
\]
so more recent observations (larger $i$) receive larger influence.

For numerical variables, the agent constructs two exponentially weighted statistics. The exponentially weighted moving average $\text{EWMA}(x)$ summarizes the recent level of the series, and the exponentially weighted volatility $\text{EWVOL}(x)$ summarizes its recent dispersion:
\[
\text{EWMA}(x) \;=\; \frac{\sum_{i=1}^{n} w_i\, x_i}{\sum_{i=1}^{n} w_i},
\qquad
\text{EWVOL}(x) \;=\; \sqrt{\frac{\sum_{i=1}^{n} w_i\, (x_i - \text{EWMA}(x))^2}{\sum_{i=1}^{n} w_i}}.
\]
Here, $\text{EWMA}(x)$ and $\text{EWVOL}(x)$ are scalar aggregations per customer and feature, computed from the weighted deviations $(x_i - \text{EWMA}(x))$ under weights $\{w_i\}_{i=1}^n$.

For categorical variables, let $c_i$ denote the category observed at record $i$ in a customer’s sequence, and let $y\in\{0,1\}$ denote the default label (1 if default occurs within 120 days of the last statement, 0 otherwise). FM Agent first estimates a category-to-risk mapping $r(\cdot)$ using the final record of each customer, where $r(v)$ is the empirical mean of $y$ for category value $v$. This produces a risk-encoded sequence $\{r(c_i)\}_{i=1}^{n}$. The agent then constructs a recency-weighted risk exposure, $\text{RiskScore}(c)$, and a recency-weighted switching intensity, $\text{EWChanges}(c)$:
\[
\text{RiskScore}(c) \;=\; \frac{\sum_{i=1}^{n} w_i\, r(c_i)}{\sum_{i=1}^{n} w_i},
\qquad
\text{EWChanges}(c) \;=\; \sum_{i=2}^{n} w_i \cdot \mathbb{I}\!\big(c_i \neq c_{i-1}\big),
\]
where $\mathbb{I}(\cdot)$ is the indicator function (equals $1$ if its condition is true and $0$ otherwise). Together, $\text{EWMA}(x)$, $\text{EWVOL}(x)$, $\text{RiskScore}(c)$, and $\text{EWChanges}(c)$ provide an automated, interpretable representation of recent behavioral levels, short-term variability, risk-aligned categorical tendencies, and temporal transition patterns, respectively, all produced by FM Agent under a fixed downstream model.

This feature set enables the downstream model to capture recent behavioral levels, short-term instability, and risk-aware categorical transition patterns, forming a fully automated and interpretable temporal representation contributed by the agent.

\begin{multicols}{2}
\begin{lstlisting}[caption={Agent-generated program implementing Feature Set 5}, label={listing:feature-set5}]
import pandas as pd
import numpy as np
from joblib import Parallel, delayed
import pyarrow.parquet as pq
import gc
from tqdm.auto import tqdm

CAT_FEATURES = [
    "B_30","B_38","D_114","D_116","D_117",
    "D_120","D_126","D_63","D_64","D_66","D_68"
]
PARALLEL_WORKERS = -1
MAX_NUMERICAL = 70
MAX_CATEGORICAL = 10
ALPHA = 0.3

def process_customer(group, num_cols, cat_cols, risk_cols):
    n = len(group)
    w = np.array([(1-ALPHA)**(n-1-i) for i in range(n)])
    f = {}
    for col in num_cols:
        v = group[col].fillna(0).values.astype(np.float32)
        ewma = np.dot(w,v)/w.sum()
        ewvol = np.sqrt(np.dot(w,(v-ewma)**2)/w.sum())
        f[f"{col}_ewma"] = ewma
        f[f"{col}_ewvol"] = ewvol
    for i,col in enumerate(cat_cols):
        r = group[risk_cols[i]].values.astype(np.float32)
        ch = (group[col]!=group[col].shift(1)).fillna(0).astype(int)
        f[f"{col}_risk_score"] = np.dot(w,r)/w.sum()
        f[f"{col}_ew_changes"] = np.dot(w,ch)
    return {"customer_ID": group["customer_ID"].iloc[0],
            **f,
            "target": group["target"].iloc[0]}

def main():
    path = "/mnt/.../train_with_labels.parquet"
    schema = pq.ParquetFile(path).schema
    all_cols = schema.names
    base_cols = ["customer_ID","S_2","target"]
    cand = [c for c in all_cols if c not in base_cols+CAT_FEATURES
            and c.startswith(("B_","D_","S_","P_","R_"))]
    num_cols = cand[:MAX_NUMERICAL]
    cat_cols = [c for c in CAT_FEATURES if c in all_cols][:MAX_CATEGORICAL]
    df = pd.read_parquet(path, columns=base_cols+num_cols+cat_cols)
    for c in num_cols: df[c] = df[c].fillna(0).astype(np.float16)
    df_last = df.sort_values(["customer_ID","S_2"]
             ).groupby("customer_ID").tail(1)
    gmean = df_last["target"].mean()
    risk_cols = []
    for c in cat_cols:
        rm = df_last.groupby(c)["target"].mean()
        rc = f"{c}_risk"
        df[rc] = df[c].map(rm).fillna(gmean).astype(np.float16)
        risk_cols.append(rc)
    df = df.sort_values(["customer_ID","S_2"])
    groups = [g for _,g in df.groupby("customer_ID")]
    res = Parallel(n_jobs=PARALLEL_WORKERS)(
        delayed(process_customer)(g,num_cols,cat_cols,risk_cols)
        for g in tqdm(groups))
    pd.DataFrame(res).to_parquet("features.parquet",index=False)

if __name__ == "__main__": 
    main()
\end{lstlisting}
\end{multicols}


\subsection*{C. Math}
In this appendix, we provide the Python code of the best programs obtained by FM agent for the three mathematical tasks in Section~\ref{subsection:math}. Their outputs reproduce the corresponding results shown in Section~\ref{subsection:math}. 

\begin{multicols}{2}
\begin{lstlisting}[caption={The best program found by FM agent for the task in Section ~\protect\ref{subsubsection:circle_packing}}, label={listing:circle_packing-best-program}]
# Your rewritten program here
# EVOLVE-BLOCK-START
import numpy as np
from scipy.optimize import linprog

def solve_radii_lp(centers):
    """
    Calculates the maximum possible radii for a given set of circle centers
    using Linear Programming. This provides the exact optimal solution.

    Args:
        centers (np.array): An array of shape (n, 2) of circle center coordinates.

    Returns:
        np.array: An array of shape (n,) containing the optimal radius for each circle.
                  Returns zeros if the solver fails.
    """
    n = centers.shape[0]
    
    # Objective function: Maximize sum(r_i), which is equivalent to
    # minimizing sum(-1 * r_i).
    c = -np.ones(n)

    # We will build the constraints for A_ub * r <= b_ub
    num_pair_constraints = n * (n - 1) // 2
    num_wall_constraints = 4 * n
    num_constraints = num_pair_constraints + num_wall_constraints
    
    A_ub = np.zeros((num_constraints, n))
    b_ub = np.zeros(num_constraints)
    
    # Constraint 1: Inter-circle non-overlap (r_i + r_j <= dist_ij)
    row = 0
    for i in range(n):
        for j in range(i + 1, n):
            dist = np.linalg.norm(centers[i] - centers[j])
            A_ub[row, i] = 1
            A_ub[row, j] = 1
            b_ub[row] = dist
            row += 1
            
    # Constraint 2: Boundary non-overlap (r_i <= dist_to_wall)
    for i in range(n):
        x, y = centers[i]
        # r_i <= x
        A_ub[row, i] = 1
        b_ub[row] = x
        row += 1
        # r_i <= 1 - x
        A_ub[row, i] = 1
        b_ub[row] = 1 - x
        row += 1
        # r_i <= y
        A_ub[row, i] = 1
        b_ub[row] = y
        row += 1
        # r_i <= 1 - y
        A_ub[row, i] = 1
        b_ub[row] = 1 - y
        row += 1
        
    # Bounds for radii: r_i >= 0
    bounds = (0, None)
    
    # Solve the linear program
    res = linprog(c, A_ub=A_ub, b_ub=b_ub, bounds=bounds, method='highs')
    
    if res.success:
        return res.x
    else:
        # Fallback in case the solver fails
        return np.zeros(n)

def construct_packing():
    """
    Constructs a high-quality arrangement of 26 circles using a three-stage
    hybrid optimization strategy.

    1.  **Geometric Initialization (The Builder):**
        - Instead of a simple grid, the circles are initialized in a staggered,
          geometrically-informed 5-6-5-6-4 hexagonal-like lattice. This provides
          a significantly better starting point that respects the principles of
          dense packing.

    2.  **Staged Physical Simulation (The Explorer + Refiner):**
        - A gradient-based Adam optimizer simulates physical forces over two phases:
          a) **Exploration Phase (~70% of iterations):** A strong, long-range
             repulsive force is applied between all circle centers. This force,
             which anneals over time, pushes circles apart to prevent premature
             clumping and helps the system discover a globally superior layout.
          b) **Refinement Phase (~30% of iterations):** The repulsive force is
             disabled, and a powerful overlap penalty is aggressively ramped up.
             This forces the circles to settle into a precise, valid, and
             tightly-packed final configuration.

    3.  **Linear Programming Polish (The Finisher):**
        - After the physical simulation finds a near-optimal set of center
          positions, the `solve_radii_lp` function is called. This LP solver
          calculates the mathematically exact maximum radii for the final centers,
          guaranteeing a perfectly valid and optimal packing for that arrangement.
    """
    n = 26
    np.random.seed(42)

    # 1. Initialization: Staggered hexagonal lattice
    centers = []
    y_step = 0.175
    y_start = (1.0 - 4 * y_step) / 2.0  # Center the pattern vertically

    # Define the structure of the 5-6-5-6-4 lattice
    row_configs = [
        (5, (np.arange(5) + 1.5) / 7.0), # Staggered relative to a 6-circle row
        (6, (np.arange(6) + 1.0) / 7.0), # Main row
        (5, (np.arange(5) + 1.5) / 7.0), # Staggered
        (6, (np.arange(6) + 1.0) / 7.0), # Main row
        (4, (np.arange(4) + 2.0) / 7.0)  # Staggered relative to the 6-circle row below
    ]

    current_y = y_start
    for _, (count, xs) in enumerate(row_configs):
        for x in xs:
            centers.append([x, current_y])
        current_y += y_step

    centers = np.array(centers)
    centers += (np.random.rand(n, 2) - 0.5) * 0.005 # Add smaller jitter

    log_radii = np.full(n, np.log(0.05))

    # 2. Adam Optimizer Parameters & State
    N_iterations = 18000
    initial_lr = 150 / 1e5
    k_initial = float(150)
    k_final = float(100000)
    repulsion_initial = 10 / 1e7
    exploration_phase_fraction = 50 / 100.0
    
    final_lr = 1e-6
    beta1, beta2, epsilon = 0.9, 0.999, 1e-8
    m_centers, v_centers = np.zeros_like(centers), np.zeros_like(centers)
    m_log_radii, v_log_radii = np.zeros_like(log_radii), np.zeros_like(log_radii)

    # Penalty annealing (starts gentler, ends stronger)
    k_ratio = (k_final / k_initial) ** (1.0 / N_iterations)
    penalty_coeff = k_initial

    # Staged repulsive force annealing
    repulsion_final = 1e-9
    exploration_phase_iterations = int(N_iterations * exploration_phase_fraction)
    repulsion_ratio = (repulsion_final / repulsion_initial) ** (1.0 / exploration_phase_iterations)
    repulsion_coeff = repulsion_initial

    # 3. Main Optimization Loop
    for t in range(1, N_iterations + 1):
        radii = np.exp(log_radii)

        # --- Forward Pass: Calculate Overlaps & Distances ---
        diffs = centers[:, np.newaxis, :] - centers[np.newaxis, :, :]
        dists_sq = np.sum(diffs**2, axis=-1)
        dists = np.sqrt(dists_sq + epsilon)
        
        radii_sums = radii[:, np.newaxis] + radii[np.newaxis, :]
        inter_circle_overlaps = np.maximum(0, radii_sums - dists)
        np.fill_diagonal(inter_circle_overlaps, 0)
        
        overlap_left = np.maximum(0, radii - centers[:, 0])
        overlap_right = np.maximum(0, centers[:, 0] + radii - 1)
        overlap_bottom = np.maximum(0, radii - centers[:, 1])
        overlap_top = np.maximum(0, centers[:, 1] + radii - 1)

        # --- Backward Pass: Calculate Gradients ---
        grad_log_radii_obj = -radii

        grad_term_inter = 2 * penalty_coeff * inter_circle_overlaps
        grad_centers_inter = np.sum(-grad_term_inter[..., np.newaxis] * (diffs / (dists[..., np.newaxis])), axis=1)
        grad_log_radii_inter = np.sum(grad_term_inter, axis=1) * radii
        
        grad_centers_boundary_x = penalty_coeff * (-2 * overlap_left + 2 * overlap_right)
        grad_centers_boundary_y = penalty_coeff * (-2 * overlap_bottom + 2 * overlap_top)
        grad_log_radii_boundary = 2 * penalty_coeff * (overlap_left + overlap_right + overlap_bottom + overlap_top) * radii

        # Repulsive force gradient (from potential energy U ~ sum(1/d))
        grad_centers_repulsion = repulsion_coeff * np.sum(-(diffs / (dists**3)[..., np.newaxis]), axis=1)

        # Total gradients
        grad_centers = grad_centers_inter + np.stack([grad_centers_boundary_x, grad_centers_boundary_y], axis=1) + grad_centers_repulsion
        grad_log_radii = grad_log_radii_obj + grad_log_radii_inter + grad_log_radii_boundary

        # --- Adam Optimizer Update with Cosine Learning Rate Decay ---
        t_ratio = t / N_iterations
        decay_factor = 0.5 * (1 + np.cos(np.pi * t_ratio))
        current_lr = final_lr + (initial_lr - final_lr) * decay_factor
        
        m_centers = beta1 * m_centers + (1 - beta1) * grad_centers
        v_centers = beta2 * v_centers + (1 - beta2) * (grad_centers**2)
        m_hat_centers = m_centers / (1 - beta1**t)
        v_hat_centers = v_centers / (1 - beta2**t)
        centers -= current_lr * m_hat_centers / (np.sqrt(v_hat_centers) + epsilon)
        
        m_log_radii = beta1 * m_log_radii + (1 - beta1) * grad_log_radii
        v_log_radii = beta2 * v_log_radii + (1 - beta2) * (grad_log_radii**2)
        m_hat_log_radii = m_log_radii / (1 - beta1**t)
        v_hat_log_radii = v_log_radii / (1 - beta2**t)
        log_radii -= current_lr * m_hat_log_radii / (np.sqrt(v_hat_log_radii) + epsilon)
        
        centers = np.clip(centers, 0.001, 0.999)
        
        # Update annealing coefficients based on the current phase
        penalty_coeff *= k_ratio
        if t < exploration_phase_iterations:
            repulsion_coeff *= repulsion_ratio
        else:
            repulsion_coeff = 0.0 # End of exploration phase, turn off repulsion

    # 4. Final Refinement with LP Solver
    final_centers = centers
    final_radii = solve_radii_lp(final_centers)
    final_sum_radii = np.sum(final_radii)
    
    return final_centers, final_radii, final_sum_radii
# EVOLVE-BLOCK-END
\end{lstlisting}
\end{multicols}

\begin{multicols}{2}
\begin{lstlisting}[caption={The best program found by FM agent for the task in Section ~\protect\ref{subsubsection:b4}}, label={listing:b4-best-program}]
# EVOLVE-BLOCK-START
"""
Hermite polynomial coefficient optimization for uncertainty inequality.

This program implements a state-of-the-art optimization strategy by combining
symbolic pre-computation within an object-oriented structure with a powerful
global optimization algorithm. This approach addresses the critical performance
and accuracy limitations of previous methods.

Key Features:
1. **Hybrid Symbolic-Numeric Class Architecture:** The core logic is encapsulated
in a class, `HermiteOptimizer`. Its constructor performs all slow, one-time
symbolic computations using `sympy`. It then "compiles" the resulting
mathematical expressions into highly efficient numerical functions using
`sympy.lambdify`, which are stored as instance attributes. This design
combines symbolic precision with numerical speed and avoids module-level
side effects.

2. **High-Performance Objective Function:** The objective function is a class
method that operates purely on numerical data using `numpy`. It leverages
the pre-compiled functions for instantaneous calculation of polynomial
coefficients and uses the fast `numpy.roots` solver. This makes each
evaluation orders of magnitude faster than a symbolic approach.

3. **Advanced Global Optimization with `differential_evolution`:** The simple
random search is replaced by `scipy.optimize.differential_evolution`, a
robust algorithm for finding global minima. It is fine-tuned with:
- **Informed Asymmetric Bounds:** The search space is focused on the most
promising region, a proven technique from past successes.
- **High-Precision Tuning:** Extremely tight tolerances (`atol=1e-15`) and
an integrated polishing step (`polish=True`) ensure the final result is
found with maximum accuracy.

4. **Correctness by Design:** The `P(0) = 0` constraint is embedded
symbolically, reducing the search space dimensionality. The physical
constraint that `P(x)` must be positive for large `|x|` is correctly
handled by checking the leading coefficient's sign.'
"""
import numpy as np
import sympy
from scipy.optimize import differential_evolution

class HermiteOptimizer:
"""
Manages the optimization process by separating one-time symbolic setup
from the fast, repetitive numerical evaluation.
"""

def __init__(self):
"""
Performs the one-time symbolic pre-computation.
"""
# 1. Define symbolic variables
x, c0, c1, c2 = sympy.symbols('x c0 c1 c2')

# 2. Define the even-order Hermite polynomials H_0, H_4, H_8, H_12
hps_sym = [
sympy.polys.orthopolys.hermite_poly(n, x=x, polys=False)
for n in [0, 4, 8, 12]
]

# 3. Symbolically compute c_3 to enforce P(0) = 0
h_vals_at_0 = [hp.subs(x, 0) for hp in hps_sym]
p_partial_at_0 = c0 * h_vals_at_0[0] + c1 * h_vals_at_0[1] + c2 * h_vals_at_0[2]
c3_expr = -p_partial_at_0 / h_vals_at_0[3]

# 4. Construct the full polynomial P(x) symbolically
P_full_expr = c0*hps_sym[0] + c1*hps_sym[1] + c2*hps_sym[2] + c3_expr*hps_sym[3]

# 5. Get the quotient polynomial gq(x) = P(x)/x^2 for root finding
gq_expr = sympy.exquo(P_full_expr, x**2)

# 6. Extract symbolic expressions for coefficients of gq(x) and P(x)
gq_poly_in_x = sympy.Poly(gq_expr, x)
P_full_poly_in_x = sympy.Poly(P_full_expr, x)

# 7. "Compile" symbolic expressions into fast numerical functions
self.gq_coeffs_func = sympy.lambdify([c0, c1, c2], gq_poly_in_x.all_coeffs(), 'numpy')
self.leading_coeff_P_func = sympy.lambdify([c0, c1, c2], P_full_poly_in_x.LC(), 'numpy')

def objective_function(self, coeffs: np.ndarray) -> float:
"""
The fast, numerical objective function for the optimizer.
"""
PENALTY = 1e12

# 1. Calculate numerical coefficients for gq(x)
try:
poly_coeffs_gq = np.array(self.gq_coeffs_func(*coeffs), dtype=float)
except (ValueError, TypeError):
return PENALTY

# 2. Enforce P(x) > 0 for large |x|
if self.leading_coeff_P_func(*coeffs) < 0:
poly_coeffs_gq = -poly_coeffs_gq

# 3. Guard against degenerate polynomials
if abs(poly_coeffs_gq[0]) < 1e-12:
return PENALTY

# 4. Find roots of gq(x) numerically
try:
roots = np.roots(poly_coeffs_gq)
except np.linalg.LinAlgError:
return PENALTY

# 5. Filter for positive real roots
real_roots = roots[np.isclose(roots.imag, 0)].real
positive_roots = real_roots[real_roots > 1e-9]

# 6. If no positive roots, A(f) = 0, the ideal global minimum
if positive_roots.size == 0:
return 0.0

# 7. Objective is to minimize r_max^2 / (2*pi)
r_max = np.max(positive_roots)
return r_max**2 / (2 * np.pi)

def run_search():
"""
Runs the optimization using a fine-tuned Differential Evolution strategy.
"""
# Instantiate the optimizer to perform the one-time symbolic setup
optimizer = HermiteOptimizer()

# Define asymmetric search bounds based on problem knowledge
bounds = [(-5.0, 5.0), (-1.0, 1.0), (-0.1, 0.1)]

# Run the Differential Evolution optimizer
result = differential_evolution(
func=optimizer.objective_function,
bounds=bounds,
strategy='best1bin',
maxiter=300,
popsize=30,
tol=1e-10,
atol=1e-15, # Critical for high-precision results
recombination=0.7,
seed=42,
polish=True,
workers=1 # Ensures compatibility and avoids pickling errors
)

best_coeffs = result.x
best_value = result.fun

return best_coeffs[0], best_coeffs[1], best_coeffs[2], best_value

# EVOLVE-BLOCK-END
if __name__ == "__main__":
print("Starting optimization search...")
coeff1, coeff2, coeff3, value = run_search()
print("\nOptimization finished.")
print(f"Found optimal coefficients: ({coeff1:.8f}, {coeff2:.8f}, {coeff3:.8f})")
print(f"Minimized upper bound value: {value:.12f}")
\end{lstlisting}
\end{multicols}

\begin{multicols}{2}
\begin{lstlisting}[caption={The best program found by FM agent for the task in Section ~\protect\ref{subsubsection:b8}}, label={listing:b8-best-program}]
"""
A high-performance program for constructing optimal 16-point configurations
in 2D space by minimizing the ratio of maximum to minimum pairwise distance.

This program solves a smooth, constrained optimization problem reformulated from the
original non-smooth ratio objective. It minimizes the maximum squared distance,
subject to the constraint that the minimum squared distance is at least 1.

Key improvements include:
- A fully vectorized Jacobian for the constraint function, providing a significant
performance boost over iterative calculations.
- Correction of a variable scope bug present in the previous version.
- An expanded multi-start strategy with an additional initial configuration to
more robustly explore the solution space.
- Increased optimizer iterations to achieve higher precision.
"""
import numpy as np
from scipy.optimize import minimize
from scipy.spatial.distance import pdist
from itertools import combinations

def calculate_ratio(points):
"""Calculate the ratio between maximum and minimum pairwise distances."""
if points is None or len(points) < 2:
return float('inf')
distances = pdist(points)
if len(distances) == 0:
return float('inf')
d_max = np.max(distances)
d_min = np.min(distances)
if d_min < 1e-9: # Treat very small distances as zero to avoid instability
return float('inf')
return d_max / d_min

def construct_16_points():
"""
Construct 16 points in 2D space to minimize the ratio d_max/d_min
using a multi-start constrained optimization approach (SLSQP) with a
highly efficient vectorized Jacobian.
"""
N_POINTS = 16
N_VARS = N_POINTS * 2
best_points = None
best_ratio_sq = float('inf')


# --- Pre-compute indices for vectorization ---
# This is done once to speed up the Jacobian calculation inside the solver loop.
indices = list(combinations(range(N_POINTS), 2))
N_PAIRS = len(indices)
I = np.array([i for i, j in indices])
J = np.array([j for i, j in indices])

# --- Initial Configurations ---
# A diverse set of starting points is crucial for finding a good global minimum.

# Config 1: Hexagonal Lattice Section
points_hex = []
sqrt3_div_2 = np.sqrt(3) / 2.0
for v_idx in range(4):
for u_idx in range(4):
x = (u_idx - 1.5) + 0.5 * (v_idx - 1.5)
y = (v_idx - 1.5) * sqrt3_div_2
points_hex.append([x, y])
initial_hex = np.array(points_hex)

# Config 2: 1-5-10 Concentric Ring Structure (known to be near-optimal)
points_1_5_10 = [[0, 0]]
r1 = 1.0
for i in range(5):
angle = i * 2 * np.pi / 5
points_1_5_10.append([r1 * np.cos(angle), r1 * np.sin(angle)])
r2 = 1.992 # Fine-tuned radius based on known good solutions
initial_rotation = np.pi / 10
for i in range(10):
angle = i * 2 * np.pi / 10 + initial_rotation
points_1_5_10.append([r2 * np.cos(angle), r2 * np.sin(angle)])
initial_1_5_10 = np.array(points_1_5_10)
# Config 3: Two-Ring Structure (6-10)
points_6_10 = []
for i in range(6):
angle = i * np.pi / 3
points_6_10.append([1.0 * np.cos(angle), 1.0 * np.sin(angle)])
for i in range(10):
angle = i * 2 * np.pi / 10 + np.pi/10
points_6_10.append([1.9 * np.cos(angle), 1.9 * np.sin(angle)])
initial_6_10 = np.array(points_6_10)

# Config 4: 4x4 Grid
points_grid = []
for i in range(4):
for j in range(4):
points_grid.append([i - 1.5, j - 1.5])
initial_grid = np.array(points_grid)

# Config 5: Random Start
np.random.seed(42)
initial_random = np.random.rand(N_POINTS, 2) * 5 - 2.5

initial_configs = {
"hexagonal": initial_hex,
"concentric_1_5_10": initial_1_5_10,
"concentric_6_10": initial_6_10,
"grid_4x4": initial_grid,
"random": initial_random,
}

# --- Setup for SLSQP Optimization ---
# The optimization variable `x` is a flat array: [x1, y1, ..., x16, y16, D_max_sq]
objective_func = lambda x: x[-1]
def objective_jac(x):
grad = np.zeros_like(x)
grad[-1] = 1.0
return grad

# --- Vectorized Constraint and Jacobian Functions ---
# Defined within this scope to have access to N_VARS, N_POINTS, etc.
def constraints_func(x):
points = x[:N_VARS].reshape(N_POINTS, 2)
D_max_sq = x[-1]
sq_dists = pdist(points, 'sqeuclidean')
# c1: d_ij^2 >= 1 => d_ij^2 - 1 >= 0
c1 = sq_dists - 1.0
# c2: d_ij^2 <= D_max_sq => D_max_sq - d_ij^2 >= 0
c2 = D_max_sq - sq_dists
return np.concatenate((c1, c2))

def constraints_jac(x):
points = x[:N_VARS].reshape(N_POINTS, 2)
jac = np.zeros((2 * N_PAIRS, N_VARS + 1))
# Calculate all 2*(pi - pj) vectors in a single operation
diffs = 2 * (points[I] - points[J])
# Row indices for the first block of constraints
k = np.arange(N_PAIRS)
# Populate Jacobian for c1 constraints (d_ij^2 - 1) using vectorized assignment
jac[k, 2 * I] = diffs[:, 0]
jac[k, 2 * I + 1] = diffs[:, 1]
jac[k, 2 * J] = -diffs[:, 0]
jac[k, 2 * J + 1] = -diffs[:, 1]
# Populate Jacobian for c2 constraints (D_max_sq - d_ij^2)
jac[k + N_PAIRS, 2 * I] = -diffs[:, 0]
jac[k + N_PAIRS, 2 * I + 1] = -diffs[:, 1]
jac[k + N_PAIRS, 2 * J] = diffs[:, 0]
jac[k + N_PAIRS, 2 * J + 1] = diffs[:, 1]
# Derivative of c2 with respect to D_max_sq is 1
jac[N_PAIRS:, -1] = 1.0
return jac

cons = {'type': 'ineq', 'fun': constraints_func, 'jac': constraints_jac}
optimizer_options = {'maxiter': 3000, 'ftol': 1e-12, 'disp': False}

for name, config in initial_configs.items():
initial_points = config.copy()
dists = pdist(initial_points)
min_dist = np.min(dists)
if min_dist > 1e-9:
initial_points /= min_dist
initial_d_max_sq = np.max(pdist(initial_points)**2)
x0 = np.append(initial_points.flatten(), initial_d_max_sq)
result = minimize(
objective_func,
x0,
method='SLSQP',
jac=objective_jac,
constraints=cons,
options=optimizer_options
)
if result.success:
current_ratio_sq = result.fun
if current_ratio_sq < best_ratio_sq:
best_ratio_sq = current_ratio_sq
best_points = result.x[:N_VARS].reshape(N_POINTS, 2)
if best_points is None:
best_points = initial_1_5_10
return best_points

def run_construction():
"""Main function that runs the construction and returns results."""
try:
points = construct_16_points()
if points is not None:
# Center and normalize the final configuration for a canonical representation
points -= np.mean(points, axis=0)
min_dist = np.min(pdist(points))
if min_dist > 1e-9:
points /= min_dist
return points
except Exception as e:
print(f"An error occurred during construction: {e}")
return None

if __name__ == "__main__":
points = run_construction()
if points is not None:
ratio = calculate_ratio(points)
ratio_squared = ratio**2
print(f"Achieved ratio squared: {ratio_squared:.20f}")
target_sq = 12.889266112
print(f"Target ratio squared: {target_sq:.20f} (ratio = {np.sqrt(target_sq):.20f})")
if ratio_squared < target_sq:
print("\nSuccess: Target beaten!")
else:
print("\nFailure: Target not beaten.")
else:
print("Construction failed.")

#@title Construction verification
import scipy as sp

if points is not None:
print(f'\nConstruction has {len(points)} points in {points.shape[1]} dimensions.')
pairwise_distances = sp.spatial.distance.pdist(points)
min_distance = np.min(pairwise_distances)
max_distance = np.max(pairwise_distances)

final_ratio_squared = (max_distance / min_distance)**2
print(f"Ratio of max distance to min distance: sqrt({final_ratio_squared:.20f})")
\end{lstlisting}
\end{multicols}

\end{document}